\documentclass[10pt,twocolumn,letterpaper]{article}

\usepackage[pagenumbers]{cvpr} 

\usepackage{graphicx}
\usepackage{amsmath}
\usepackage{amssymb}
\usepackage{booktabs}
\usepackage{xcolor}         
\usepackage{dblfloatfix}
\usepackage[accsupp]{axessibility} 

\usepackage[pagebackref,breaklinks,colorlinks]{hyperref}

\newcommand{\mill}{Mill 19}
\newcommand{\meganerf}{Mega-NeRF}
\newcommand{\meganerffast}{Mega-NeRF-Dynamic}

\DeclareMathOperator*{\argmin}{argmin}

\usepackage[capitalize]{cleveref}
\crefname{section}{Sec.}{Secs.}
\Crefname{section}{Section}{Sections}
\Crefname{table}{Table}{Tables}
\crefname{table}{Tab.}{Tabs.}

\def\cvprPaperID{9405} 
\def\confName{CVPR}
\def\confYear{2022}

\begin{document}

\title{\meganerf: \\Scalable Construction of Large-Scale NeRFs for Virtual Fly-Throughs}

\author{Haithem Turki\textsuperscript{1} \qquad Deva Ramanan\textsuperscript{1,2} \qquad Mahadev Satyanarayanan\textsuperscript{1}
\\
\textsuperscript{1}Carnegie Mellon University \qquad \textsuperscript{3}Argo AI } 

\maketitle

\begin{abstract}
   We use neural radiance fields (NeRFs) to build interactive 3D environments from large-scale visual captures spanning buildings or even multiple city blocks collected primarily from drones. In contrast to single object scenes (on which NeRFs are traditionally evaluated), our scale poses multiple challenges including (1) the need to model thousands of images with varying lighting conditions, each of which capture only a small subset of the scene, (2) prohibitively large model capacities that make it infeasible to train on a single GPU, and (3) significant challenges for fast rendering that would enable interactive fly-throughs.
   To address these challenges, we begin by analyzing visibility statistics for large-scale scenes, motivating a sparse network structure where parameters are specialized to different regions of the scene. We introduce a simple geometric clustering algorithm for data parallelism that partitions training images (or rather pixels) into different NeRF submodules that can be trained in parallel. 
   We evaluate our approach on existing datasets (Quad 6k and UrbanScene3D) as well as against our own drone footage, improving training speed by 3x and PSNR by 12\%. We also evaluate recent NeRF fast renderers on top of \meganerf\ and introduce a novel method that exploits temporal coherence. Our technique achieves a 40x speedup over conventional NeRF rendering while remaining within 0.8 db in PSNR quality, exceeding the fidelity of existing fast renderers.
\end{abstract}
\vspace*{-5mm}

\section{Introduction}
\label{sec:intro}

\begin{figure}
    \centering
    \begin{tabular}[t]{cc}
\begin{subfigure}{0.65\linewidth}
    \centering
    \includegraphics[width=0.9\linewidth, clip=true,trim = 20mm 0mm 0mm 0mm 0mm]{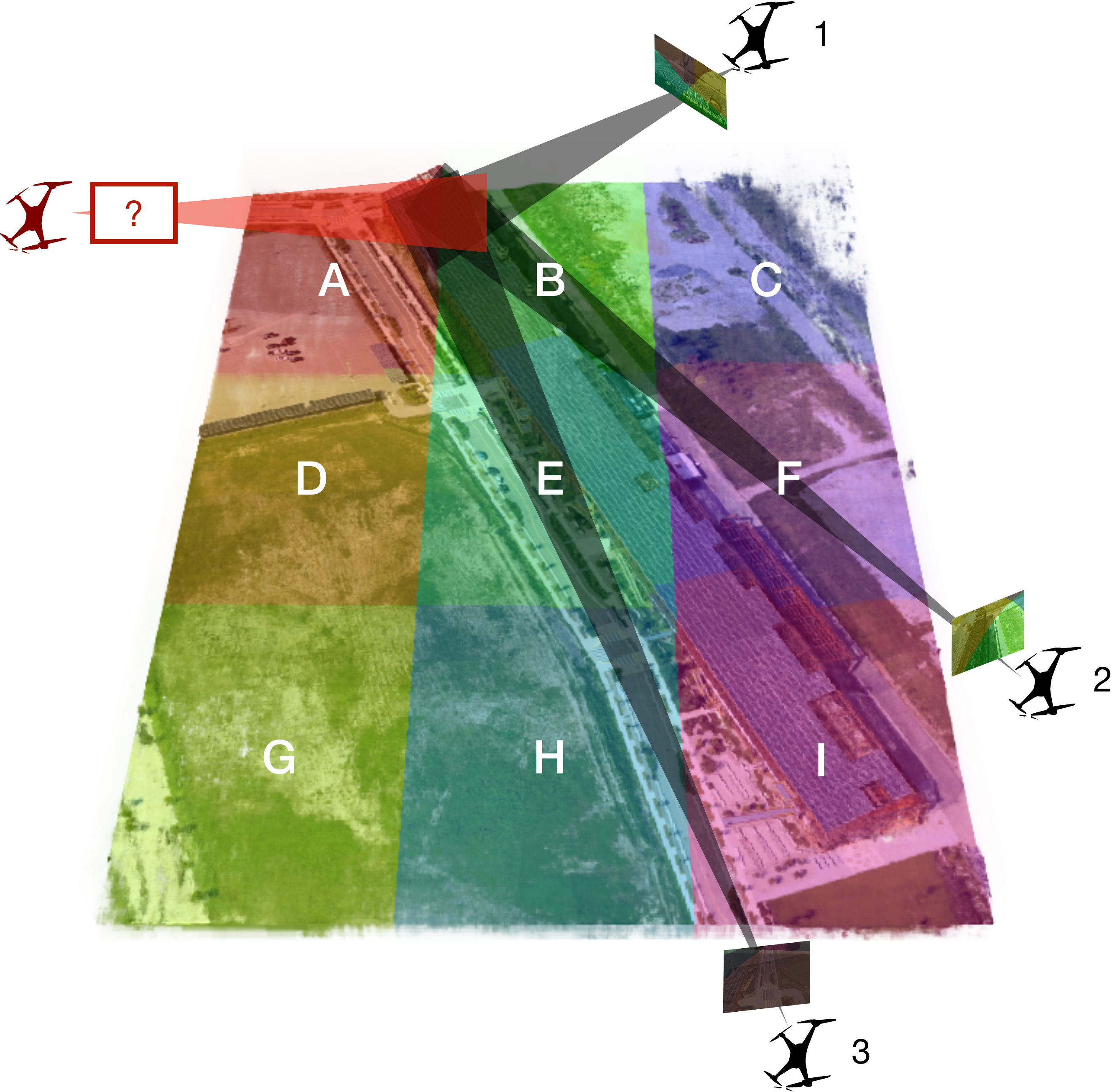}
\end{subfigure}
\hspace*{-10mm}
    &
        \begin{tabular}{c}
        \smallskip
            \begin{subfigure}[t]{0.4\linewidth}
                \centering
                \includegraphics[width=0.9\linewidth]{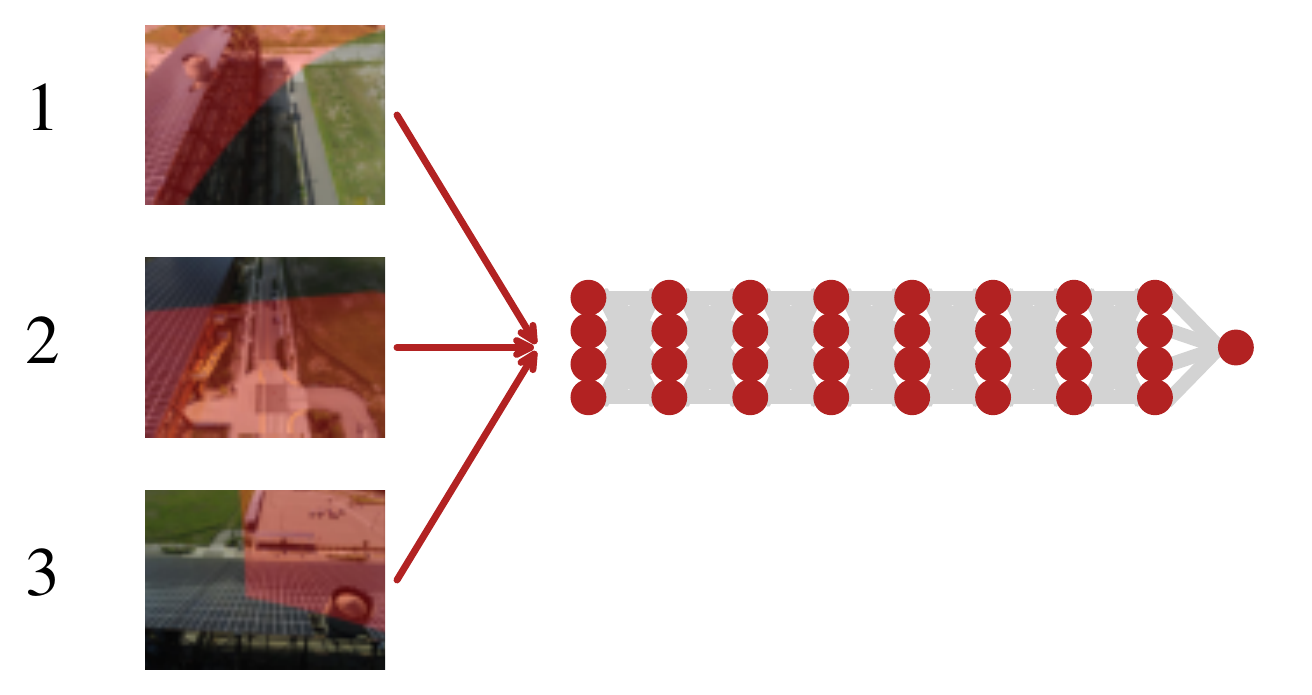}
                \vspace*{-1mm}
                \caption*{Training: Data Partitioning}
                \vspace*{1mm}
            \end{subfigure}\\
            \begin{subfigure}[t]{0.4\linewidth}
                \centering
                \includegraphics[width=0.9\textwidth]{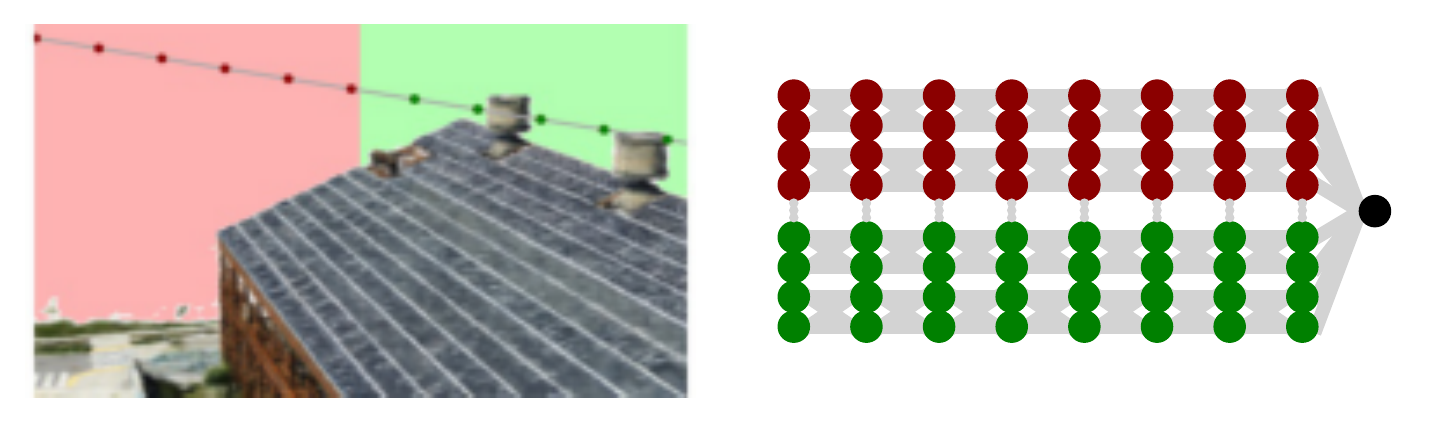}
                \vspace*{-1mm}
                \caption*{Inference: View Synthesis}
            \end{subfigure}
        \end{tabular}\\
    \end{tabular}
\caption{We scale neural reconstructions to massive urban scenes 1000x larger than prior work. To do so, \meganerf\ decomposes a scene into a set of spatial cells (\textbf{left}), learning a separate NeRF submodule for each. We train each submodule with geometry-aware pixel-data partitioning, making use of {\em only} those pixels whose rays intersect that spatial cell (\textbf{top right}). For example, pixels from image 2 are added to the trainset of cells A, B, and F, reducing the size of each trainset by 10x. To generate new views for virtual fly-throughs, we make use of standard raycasting and point sampling, but query the encompassing submodule for each sampled point (\textbf{bottom right}). To ensure view generation is near-interactive, we make use of temporal coherence by caching occupancy and color values from nearby previous views (Fig.~\ref{fig:voxel-refinement}).}
\label{fig:teaser}
\vspace*{-7mm}
\end{figure}

Recent advances in neural rendering techniques have lead to significant progress towards photo-realistic novel view synthesis, a prerequisite towards many VR and AR applications. In particular, Neural Radiance Fields (NeRFs)~\cite{mildenhall2020nerf} have attracted significant attention, spawning a wide range of follow-up works that improve upon various aspects of the original methodology.

{\bf Scale.} Simply put, our work explores the scalability of NeRFs. The vast majority of existing methods explore single-object scenes, often captured indoors or from synthetic data. To our knowledge, Tanks and Temples~\cite{Knapitsch2017} is the largest dataset used in NeRF evaluation, spanning 463 $m^2$ on average. In this work, we scale NeRFs to capture and interactively visualize urban-scale environments from drone footage that is orders of magnitude larger than any dataset to date, from 150,000 to over 1,300,000 $m^2$ per scene.

{\bf Search and Rescue.} As a motivating use case, consider search-and-rescue, where drones provide an inexpensive means of quickly surveying an area and prioritizing limited first responder resources (e.g., for ground team deployment). Because battery life and bandwidth limits the ability to capture sufficiently detailed footage in real-time~\cite{djiwhitepaper}, collected footage is typically reconstructed into 2D ``birds-eye-view" maps that support post-hoc analysis~\cite{drones4030038}. We imagine a future in which neural rendering lifts this analysis into 3D, enabling response teams to inspect the field as if they were flying a drone in real-time at a level of detail far beyond the achievable with classic Structure-from-Motion (SfM).

{\bf Challenges.} Within this setting, we encounter multiple challenges. Firstly, applications such as search-and-rescue are time-sensitive. According to the National Search and Rescue Plan~\cite{SAR}, “the life expectancy of an injured survivor decreases as much as 80 percent during the first 24 hours, while the chances of survival of uninjured survivors rapidly diminishes after the first 3 days.” The ability to train a usable model within a few hours would therefore be highly valuable. Secondly, as our datasets are orders of magnitude larger than previously evaluated datasets (Table~\ref{table:dataset-statistics}), model capacity must be significantly increased in order to ensure high visual fidelity, further increasing training time. Finally, although interactive rendering is important for fly-through and exploration at the scale we capture, existing real-time NeRF renderers either rely on pretabulating outputs into a finite-resolution structure, which scales poorly and significantly degrades rendering performance, or require excessive preprocessing time.

\begin{table}
\resizebox{\linewidth}{!}{
\begin{tabular}{l@{\hspace{1em}}c@{\hspace{1em}}c@{\hspace{1em}}c@{\hspace{1em}}c}
\toprule 
&  & &   & Scene Captured \\
&  Resolution & \# Images & \# Pixels/Rays & / Image \\
Synthetic NeRF - Chair & 400 x 400 & 400 & 256,000,000 & 0.271  \\
Synthetic NeRF - Drums & 400 x 400 & 400 & 256,000,000 & 0.302  \\
Synthetic NeRF - Ficus & 400 x 400 & 400 & 256,000,000 & 0.582  \\
Synthetic NeRF - Hotdog & 400 x 400 & 400 & 256,000,000 & 0.375  \\
Synthetic NeRF - Lego & 400 x 400 & 400 & 256,000,000 & 0.205  \\
Synthetic NeRF - Materials & 400 x 400 & 400 & 256,000,000 & 0.379  \\
Synthetic NeRF - Mic & 400 x 400 & 400 & 256,000,000 & 0.518  \\
Synthetic NeRF - Ship & 400 x 400 & 400 & 256,000,000 & 0.483 \\
T\&T - Barn & 1920 x 1080 & 384 & 796,262,400 & 0.135 \\
T\&T - Caterpillar & 1920 x 1080 & 368 & 763,084,800 & 0.216 \\
T\&T - Family & 1920 x 1080 & 152 & 315,187,200 & 0.284 \\
T\&T - Ignatius & 1920 x 1080 & 263 & 545,356,800 & 0.476 \\
T\&T - Truck & 1920 x 1080 & 250 & 518,400,000 & 0.225 \\
\midrule
\mill\ - Building & 4608 x 3456  & 1940 & 30,894,981,120 & 0.062 \\
\mill\ - Rubble & 4608 x 3456 & 1678 & 26,722,566,144 & 0.050 \\
Quad 6k &  1708 x 1329 & 5147 & 11,574,265,679 & 0.010 \\
UrbanScene3D - Residence & 5472 x 3648 & 2582 & 51,541,512,192 & 0.059 \\
UrbanScene3D - Sci-Art  & 4864 x 3648 & 3019 & 53,568,749,568 & 0.088 \\
UrbanScene3D - Campus   & 5472 x 3648 & 5871 & 117,196,056,576 & 0.028 \\
\bottomrule
\end{tabular}
}
\caption{Scene properties from the commonly used Synthetic NeRF and Tanks and Temples datasets (T\&T) compared to our target datasets ({\bf below}). Our targets contain an order-of-magnitude more pixels (and hence rays) than prior work. Moreoever, each image captures significantly less of the scene, motivating a modular approach where spatially-localized submodules are trained with a fraction of relevant image data. We provide more details and additional statistics in Sec.~\ref{sec:more-statistics} of the supplement.}
\vspace*{-5mm}
\label{table:dataset-statistics}
\end{table}

{\bf \meganerf.} In order to address these issues, we propose \meganerf, a framework for training large-scale 3D scenes that support interactive human-in-the-loop fly-throughs. We begin by analyzing visibility statistics for large-scale scenes, as shown in Table~\ref{table:dataset-statistics}. Because only a small fraction of the training images are visible from any particular scene point, we introduce a sparse network structure where parameters are specialized to different regions of the scene. We introduce a simple geometric clustering algorithm that partitions training images (or rather pixels) into different NeRF submodules that can be trained in parallel. We further exploit spatial locality at render time to implement a just-in-time visualization technique that allows for interactive fly-throughs of the captured environment. 

{\bf Prior art.} Our approach of using ``multiple" NeRF submodules is closely inspired by the recent work of DeRF~\cite{9578215} and KiloNeRF~\cite{reiser2021kilonerf}, which use similar insights to accelerate {\em inference} (or rendering) of an existing, pre-trained NeRF. However, even obtaining a pre-trained NeRF for our scene scales is essentially impossible with current training pipelines. We demonstrate that modularity is vital for {\em training}, particularly when combined with an intelligent strategy for ``sharding" training data into the appropriate modules via geometric clustering.

{\bf Contributions.} We propose a reformulation of the NeRF architecture that sparsifies layer connections in a spatially-aware manner, facilitating efficiency improvements at training and rendering time. We then adapt the training process to exploit spatial locality and train the model subweights in a fully parallelizable manner, leading to a 3x improvement in training speed while exceeding the reconstruction quality of existing approaches. In conjunction, we evaluate existing fast rendering approaches against our trained \meganerf\ model and present a novel method that exploits temporal coherence. Our technique requires minimal preprocessing, avoids the finite resolution shortfalls of other renderers, and maintains a high level of visual fidelity. We also present a new large-scale dataset containing thousands of HD images gathered from drone footage over 100,000 $m^2$ of terrain near an industrial complex.

\begin{figure}
        \centering
        \includegraphics[width=\linewidth]{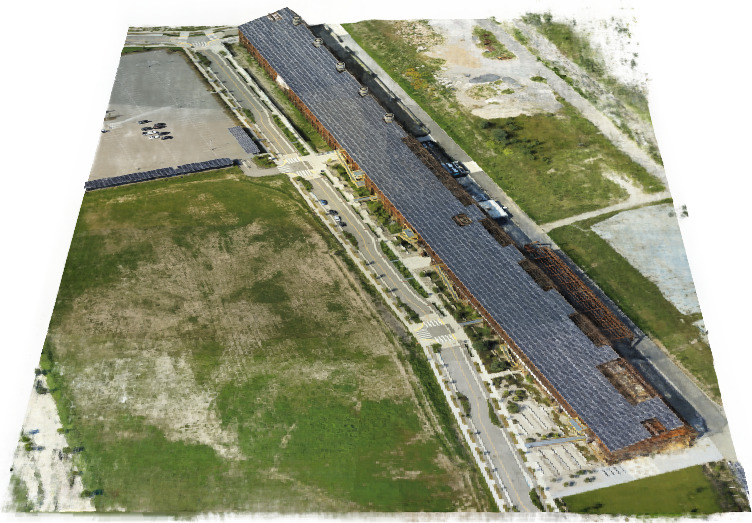}
        \includegraphics[width=\linewidth]{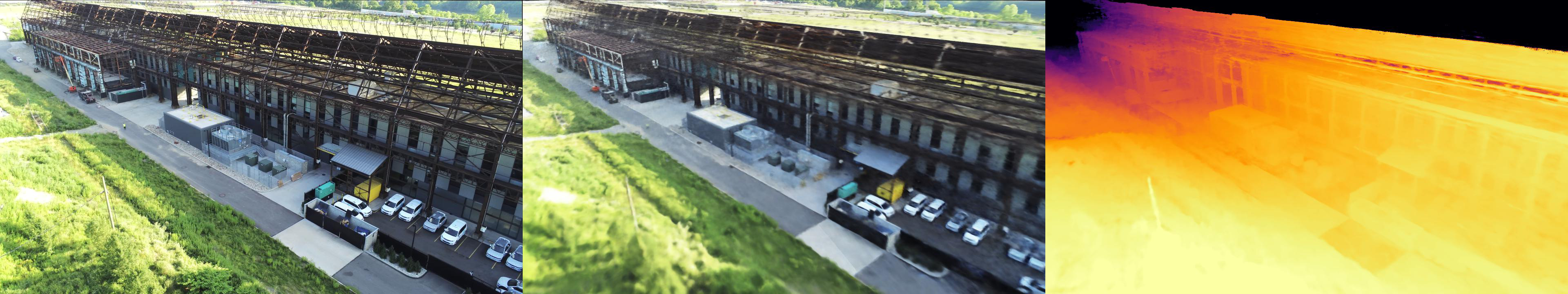}
   \caption{Visualization of \mill\ by \meganerf. The top panel shows a high-level 3D rendering of \mill\ within our interactive visualizer. The bottom-left panel contains a ground truth image captured by our drone. The following two panels illustrate the model reconstruction along with the associated depth map.}
\label{fig:mill19viz}
\vspace*{-5mm}
\end{figure}

\section{Related work}
\label{sec:related-work}

\textbf{Fast rendering.} Conventional NeRF rendering falls well below interactive thresholds. Plenoctree~\cite{yu2021plenoctrees}, SNeRG~\cite{hedman2021snerg}, and FastNeRF~\cite{garbin2021fastnerf} speed up the process by storing precomputed non-view dependent model outputs into a separate data structure such as a sparse voxel octree. These renderers then bypass the original model entirely at render time by computing the final view-dependent radiance through a separate smaller multi-layer perceptron (MLP) or through spherical basis computation. Although they achieve interactivity, they suffer from the finite capacity of the caching structure and poorly capture low-level details at scale.

DeRF~\cite{9578215} decomposes the scene into multiple cells via spatial Voronoi partitioning. Each cell is independently rendered using a smaller MLP, accelerating rendering by 3x over NeRF. KiloNeRF~\cite{reiser2021kilonerf} divides the scene into thousands of even smaller networks. Although similar in spirit to \meganerf, these methods use spatial partitioning to speed up \textit{inference} while we use it to enable \textit{data parallelism} for scalable training. Both DeRF and KiloNERF are initialized with a single large network trained on all data which is then distilled into smaller networks for fast inference, increasing processing time by over 2x for KiloNeRF. Training on all available data is prohibitive at our scale. Instead, our crucial insight is to geometrically partition training pixels into small data shards relevant for each submodule, which is essential for efficient training and high accuracy.

DONeRF~\cite{neff2021donerf} accelerates rendering by significantly reducing the number of samples queried per ray. To maintain quality, these samples are placed more closely around the first surface the ray intersects, similar to our guided sampling approach described in Sec.~\ref{subsection:rendering}. In contrast to our method, DONeRF uses a separate depth oracle network trained against ground truth depth data.

\textbf{Unbounded scenes.} Although most NeRF-related work targets indoor areas, NeRF++~\cite{zhang2020npp} handles unbounded environments by partitioning the space into a unit sphere foreground region that encloses all camera poses and a background region that covers the inverted sphere complement. A separate MLP model represents each area and performs ray casting independently before a final composition. \meganerf\ employs a similar foreground/background partitioning although we further constrain our foreground and sampling bounds as described in Sec.~\ref{subsection:model-architecture}.

NeRF in the Wild~\cite{martinbrualla2020nerfw} augments NeRF's model with an additional transient head and learned per-image embeddings to better explain lighting differences and transient occlusions across images. Although it does not explicitly target unbounded scenes, it achieves impressive results against outdoor sequences in the Phototourism~\cite{phototourism} dataset. We adopt similar appearance embeddings for \meganerf\ and quantify its impact in Sec.~\ref{subsection:scalable-training}. 

Concurrent to us, Urban Radiance Fields~\cite{rematas2022urf} (URF), CityNeRF~\cite{xiangli2021citynerf}, and BlockNeRF~\cite{tancik2022blocknerf} target urban-scale environments. URF makes use of lidar inputs, while CityNeRF makes use of multi-scale data modeling. Both methods can be seen as complementary to our approach, implying combining them with \meganerf\ is promising. Most related to us is BlockNeRF~\cite{tancik2022blocknerf}, which decomposes a scene into spatial cells of fixed city blocks. \meganerf\ makes use of geometry visibility reasoning to decompose the set of training pixels, allowing for pixels captured from far-away cameras to still influence a spatial cell (Fig.~\ref{fig:teaser}).

\textbf{Training speed.} Several works speed up model training by incorporating priors learned from similar datasets. PixelNeRF~\cite{9577688}, IBRNet~\cite{wang2021ibrnet}, and GRF~\cite{grf2020} condition NeRF on predicted image features while Tancik et al. \cite{tancik2020meta} use meta-learning to find good initial weight parameters that converge quickly. We view these efforts as complementary to ours.

\textbf{Graphics.} We note longstanding efforts within the graphics community covering interactive walkthroughs. Similar to our spatial partioning, Teller and S\'{e}quin~\cite{10.1145/122718.122725} subdivide a scene into cells to filter out irrelevant geometry and speed up rendering. Funkhouser and S\'{e}quin~\cite{10.1145/166117.166149} separately describe an adaptive display algorithm that iteratively adjusts image quality to achieve interactive frame rates within complex virtual environments. Our renderer takes inspiration from this gradual refinement approach.

\textbf{Large-scale SfM.} We take inspiration from previous large-scale reconstruction efforts based on classical Structure-from-Motion (SfM), in particular Agarwal et al's seminal ``Building Rome in a Day,"~\cite{10.1145/2001269.2001293} which describes city-scale 3D reconstruction from internet-gathered data.

\section{Approach}

We first describe our model architecture in Sec.~\ref{subsection:model-architecture}, then our training process in \ref{subsection:training}, and finally propose a novel renderer that exploits temporal coherence in~\ref{subsection:rendering}. 

\subsection{Model Architecture}
\label{subsection:model-architecture}

\textbf{Background.} We begin with a brief description of Neural Radiance Fields (NeRFs)~\cite{mildenhall2020nerf}. NeRFs represent a scene within a continuous volumetric radiance field that captures both geometry and view-dependent appearance. NeRF encodes the scenes within the weights of a multilayer perceptron (MLP). At render time, NeRF projects a camera ray $\textbf{r}$ for each image pixel and samples along the ray. For a given point sample $p_i$, NeRF queries the MLP at position $\textbf{x}_i = (x, y, z)$ and ray viewing direction $\textbf{d} = (d_1, d_2, d_3)$ to obtain opacity and color values $\sigma_i$ and $\textbf{c}_i = (r, g, b)$. It then composites a color prediction $\hat{C}(\textbf{r})$ for the ray using numerical quadrature $\sum_{i=0}^{N-1} T_i (1 - \exp( -\sigma_{i} \delta_{i})) \, \textbf{c}_i$, where $T_i = \exp( -\sum_{j=0}^{i-1} \sigma_j \delta_j)$ and $\delta_i$ is the distance between samples $p_i$ and $p_{i+1}$. The training process optimizes the model by sampling batches $R$ of image pixels and minimizing the loss function $\sum_{\textbf{r} \in \mathcal{R}} \big\lVert{C(\textbf{r}) - \hat{C}(\textbf{r})}\big\rVert^2$. NeRF samples camera rays through a two-stage hierarchical sampling process and uses positional encoding to better capture high-frequency details. We refer the reader to the NeRF paper~\cite{mildenhall2020nerf} for additional information.

\textbf{Spatial partitioning.} \meganerf\ decomposes a scene into cells with centroids $\textbf{n}_{\in \mathcal{N}} = (n_x, n_y, n_z)$ and initializes a corresponding set of model weights $f^\textbf{n}$. Each weight submodule is a sequence of fully connected layers similar to the NeRF architecture. Similar to NeRF in the Wild~\cite{martinbrualla2020nerfw}, we associate an additional appearance embedding vector $l^{(a)}$ for each input image $a$ used to compute radiance. This allows \meganerf\ additional flexibility in explaining lighting differences across images which we found to be significant at the scale of the scenes that we cover. At query time, \meganerf\ produces an opacity $\sigma$ and color $\textbf{c} = (r, g, b)$ for a given position $\textbf{x}$, direction $\textbf{d}$, and appearance embedding $l^{(a)}$ using the model weights $f^\textbf{n}$ closest to the query point:

\begin{align}
    &f^\textbf{n}(\textbf{x})  = \sigma \\
    &f^\textbf{n}(\textbf{x}, \textbf{d}, l^{(a)}) = \textbf{c} \\
     & \text{where}\ \textbf{n} = \argmin_{n \in \mathcal{N}} \big\lVert{n - \textbf{x}}\big\rVert^2
\end{align}

\textbf{Centroid selection.} Although we explored several methods, including k-means clustering and uncertainty-based partitioning as in \cite{DBLP:journals/corr/abs-2105-09103}, we ultimately found that tessellating the scene into a top-down 2D grid worked well in practice. This method is simple to implement, requires minimal preprocessing, and enables efficient assignment of point queries to centroids at inference time. As the variance in altitude between camera poses in our scenes is small relative to the differences in latitude and longitude, we fix the height of the centroids to the same value.

\begin{figure}[t!]
\centering
\includegraphics[width=.5\linewidth,clip=true,trim=0mm 6mm 0mm 18mm ]{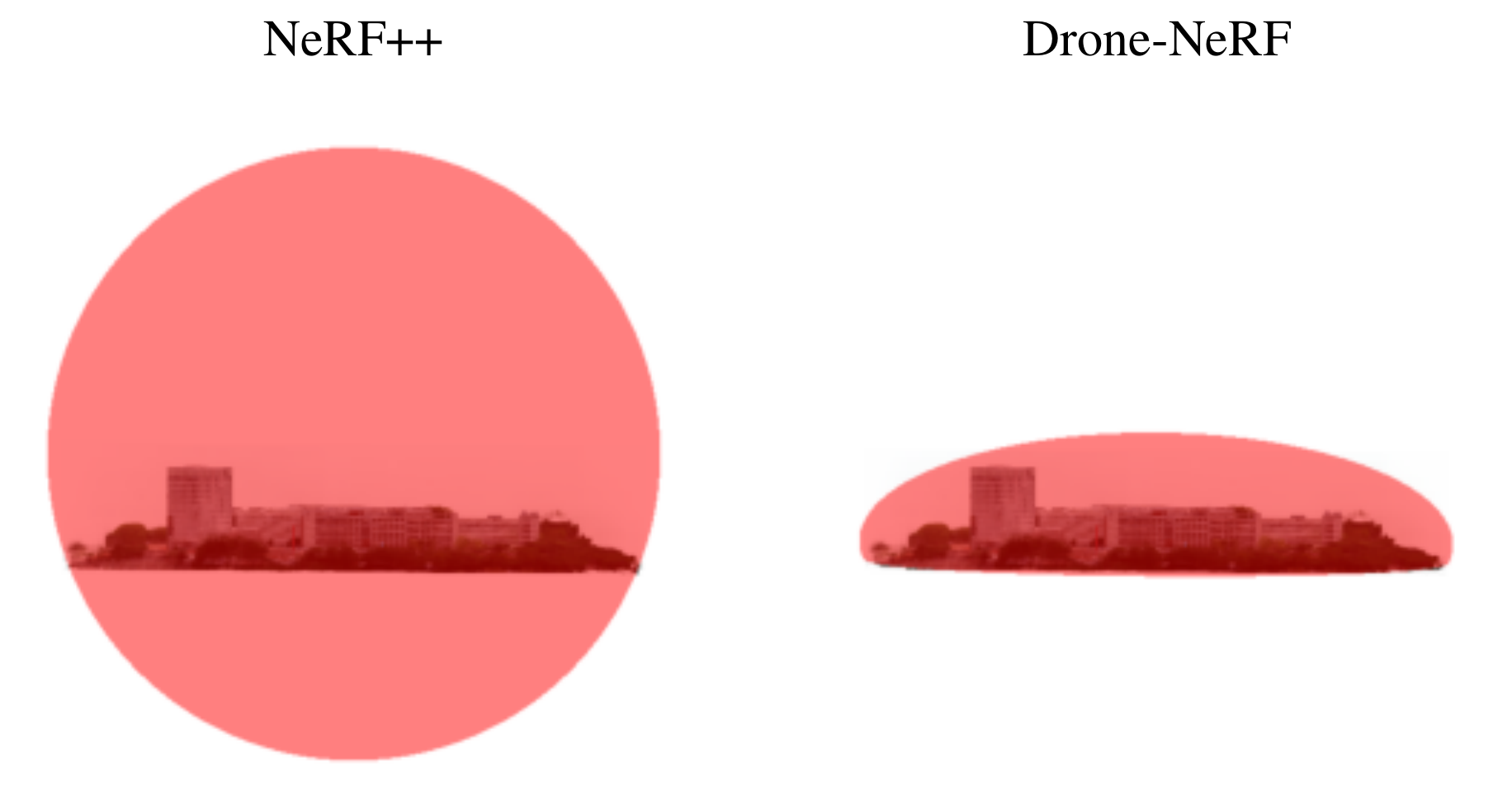}
   \caption{{\bf Ray Bounds.} NeRF++ ({\bf left}) samples within a unit sphere centered within and enclosing all camera poses to render its foreground component and uses a different methodology for the outer volume complement to efficiently render the background. \meganerf\ ({\bf right}) uses a similar background parameterization but models the foreground as an ellipsoid to achieve tighter bounds on the region of interest. It also uses camera altitude measurements to constrain ray sampling and not query underground regions.}
\label{fig:workspace-bounds}
\vspace*{-5mm}
\end{figure}

\textbf{Foreground and background decomposition.} Similar to NeRF++~\cite{zhang2020npp}, we further subdivide the scene into a foreground volume enclosing all camera poses and a background covering the complementary area. Both volumes are modeled with separate \meganerf s. We use the same 4D outer volume parameterization and raycasting formulation as NeRF++ but improve upon its unit sphere partitioning by instead using an ellipsoid that more tightly encloses the camera poses and relevant foreground detail. We also take advantage of camera altitude measurements to further refine the sampling bounds of the scene by terminating rays near ground level. \meganerf\ thus avoids needlessly querying underground regions and samples more efficiently. Fig.~\ref{fig:workspace-bounds} illustrates the differences between both approaches.

\subsection{Training}
\label{subsection:training}

\textbf{Spatial Data Parallelism.} As each \meganerf\ submodule is a self-contained MLP, we can train each in parallel with no inter-module communication. Crucially, as each image captures only a small part of the scene (Table~\ref{table:dataset-statistics}), we limit the size of each submodule's trainset to only those potentially relevant pixels. Specifically, we sample points along the camera ray corresponding to each pixel for each training image, and add that pixel to the trainset for only those spatial cells it intersects (Fig.~\ref{fig:teaser}). In our experiments, this visibility partitioning reduces the size of each submodule's trainset by 10x compared to the initial aggregate trainset. This data reduction should be even more extreme for larger-scale scenes; when training a NeRF for North Pittsburgh, one need not add pixels of South Pittsburgh. We include a small overlap factor between cells (15\% in our experiments) to further minimize visual artifacts near boundaries.

\textbf{Spatial Data Pruning.} Note that the initial assignment of pixels to spatial cells is based on camera positions, irrespective of scene geometry (because that is not known at initialization). Once NeRF gains a coarse understanding of the scene, one could further prune away irrelevant pixels/rays that don't contribute to a particular NeRF due to an intervening occluder. For example, in Fig.~\ref{fig:teaser}, early NeRF optimization might infer a wall in cell F, implying that pixels from image 2 can then be pruned from cell A and B. Our initial exploration found that this additional visibility pruning further reduced trainset sizes by 2x. We provide details in Sec.~\ref{sec:data-pruning} of the supplement.

\begin{figure*}[t!]
\centering
\includegraphics[width=\textwidth, clip = true, trim = 0mm 2mm 0mm 0mm]{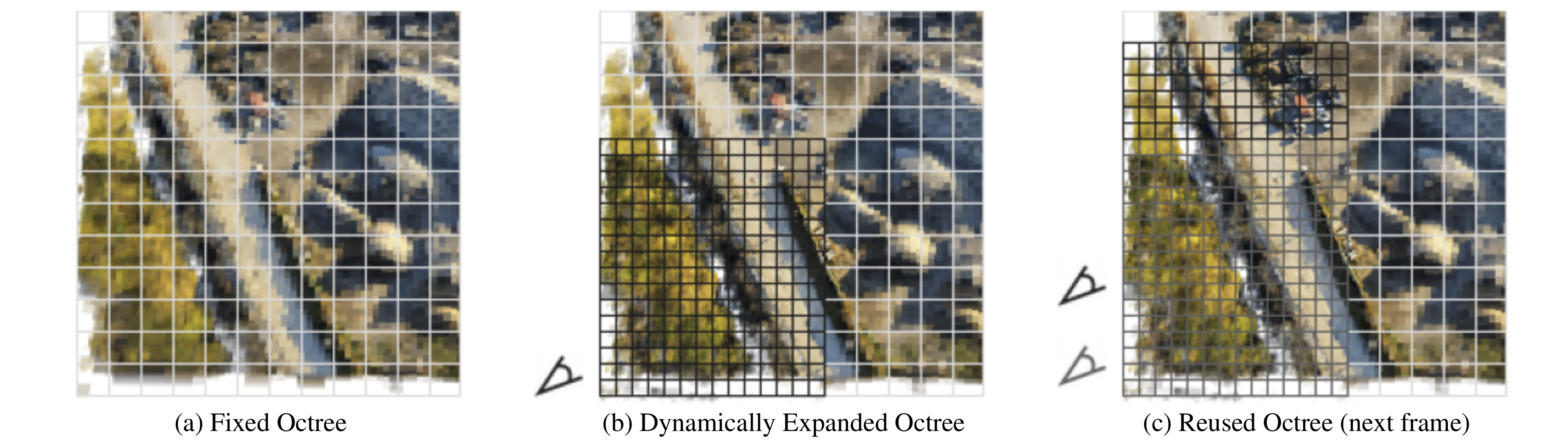}
\vspace*{-5mm}
   \caption{{\bf \meganerffast.} Current renderers (such as Plenoctree~\cite{yu2021plenoctrees}) cache precomputed model outputs into a fixed octree, limiting the resolution of rendered images ({\bf a}). \meganerffast\ {\em dynamically} expands the octree based on the current position of the fly-through ({\bf b}). Because of the temporal coherence of camera views, the next-frame rendering ({\bf c}) can reuse of much of expanded octree.}
\label{fig:voxel-refinement}
\vspace*{-5mm}
\end{figure*}

\subsection{Interactive Rendering}
\label{subsection:rendering}

We propose a novel interactive rendering method in addition to an empirical evaluation of existing fast renderers on top of \meganerf\ in Sec.~\ref{subsection:interactive-exploration}. In order to satisfy our search-and-rescue usecase, we attempt to: (a) preserve visual fidelity, (b) minimize any additional processing time beyond training the base model, and (c) accelerate rendering, which takes over 2 minutes for a 720p frame with normal ray sampling, to something more manageable.

\textbf{Caching.} Most existing fast NeRF renderers make use of cached precomputation to speed up rendering, which may not be effective at our scene scale. For example, Plenoctree~\cite{yu2021plenoctrees} precomputes a cache of opacity and spherical harmonic coefficients into a sparse voxel octree. Generating the entire 8-level octree for our scenes took an hour of computation and anywhere from 1 to 12 GB of memory depending on the radiance format. Adding a single additional level increased the processing time to 10 hours and the octree size to 55GB, beyond the capacity of all but the largest GPUs. 

\textbf{Temporal coherence.}
We explore an orthogonal direction that exploits the temporal coherence of interactive fly-throughs; once the information needed to render a given view is computed, we reuse much of it for the {\em next} view. Similar to Plenoctree, we begin by precomputing a coarse cache of opacity and color.
In contrast to Plenoctree, we {\em dynamically} subdivide the tree throughout the interactive visualization. Fig.~\ref{fig:voxel-refinement} illustrates our approach. As the camera traverses the scene, our renderer uses the cached outputs to quickly produce an initial view and then performs additional rounds of model sampling to further refine the image, storing these new values into the cache. As each subsequent frame has significant overlap with its predecessor, it benefits from the previous refinement and needs to only perform a small amount of incremental work to maintain quality. We provide further details in Sec.~\ref{sec:octree-generation} of the supplement.

\begin{figure}[t!]
\centering
\includegraphics[width=\linewidth, clip=true, trim=0mm 2.5mm 0mm 11.8mm]{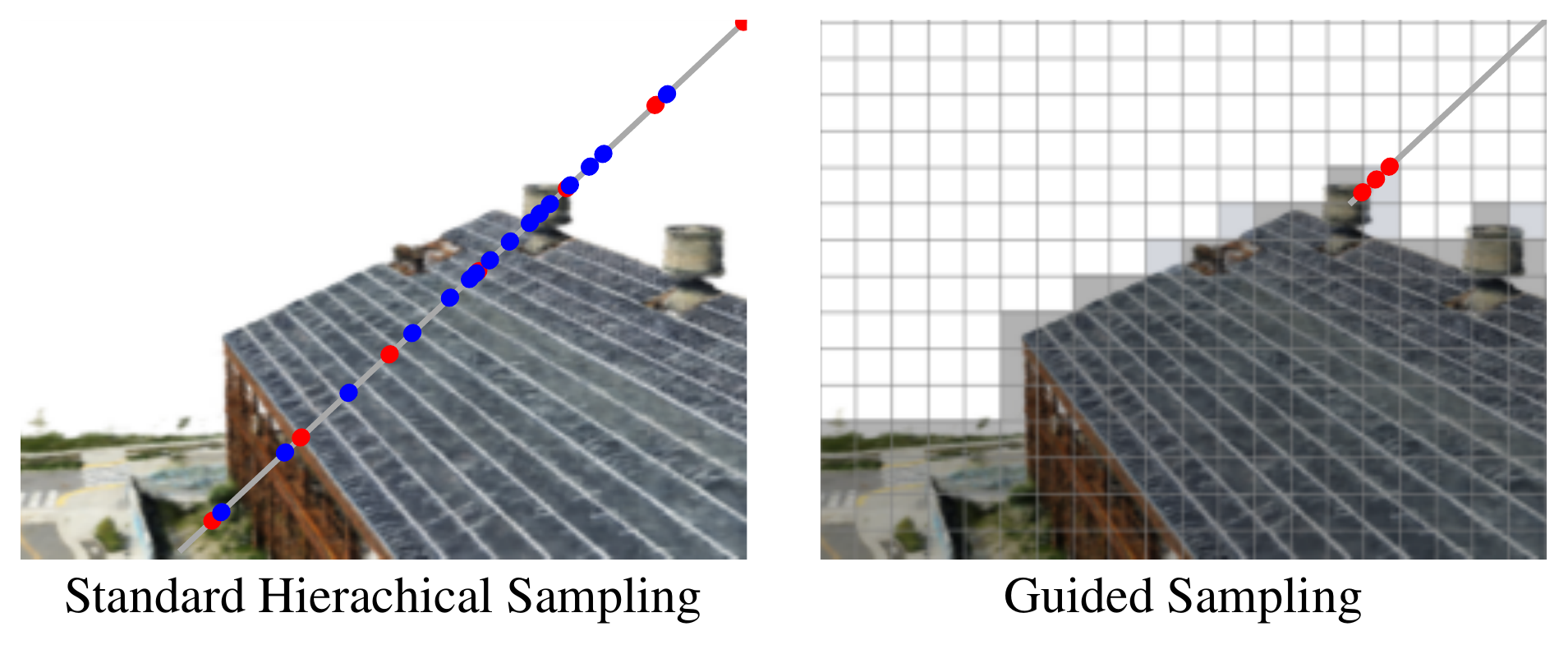}
  \caption{{\bf Guided Sampling.}
  Standard NeRF \textbf{(left)} first samples coarsely at uniform intervals along the ray and subsequently performs another round of sampling guided by the coarse weights. \meganerffast\ \textbf{(right)} uses its caching structure to skip empty spaces and take a small number of samples near surfaces.}
  \vspace*{-5mm}
\label{fig:accelerated-rendering}
\end{figure}

\textbf{Guided sampling.} We perform a final round of guided ray sampling after refining the octree to further improve rendering quality. We render rays in a single pass in contrast to NeRF's traditional two-stage hierachical sampling by using the weights stored in the octree structure. As our refined octree gives us a high-quality estimate of the scene geometry, we need to place only a small number of samples near surfaces of interest. Fig.~\ref{fig:accelerated-rendering} illustrates the difference between both approaches. Similar to other fast renderers, we further accelerate the process by accumulating transmittance along the ray and ending sampling after a certain threshold. 

\section{Experiments}

Our evaluation of \meganerf\ is motivated by the following two questions. First, given a finite training budget, how accurately can \meganerf\ capture a scene? Furthermore, after training, is it possible to render accurately at scale while minimizing latency?

\begin{figure*}[!htbp]
\centering
\includegraphics[width=\linewidth, clip = true, trim = 0mm 3mm 0mm 0mm]{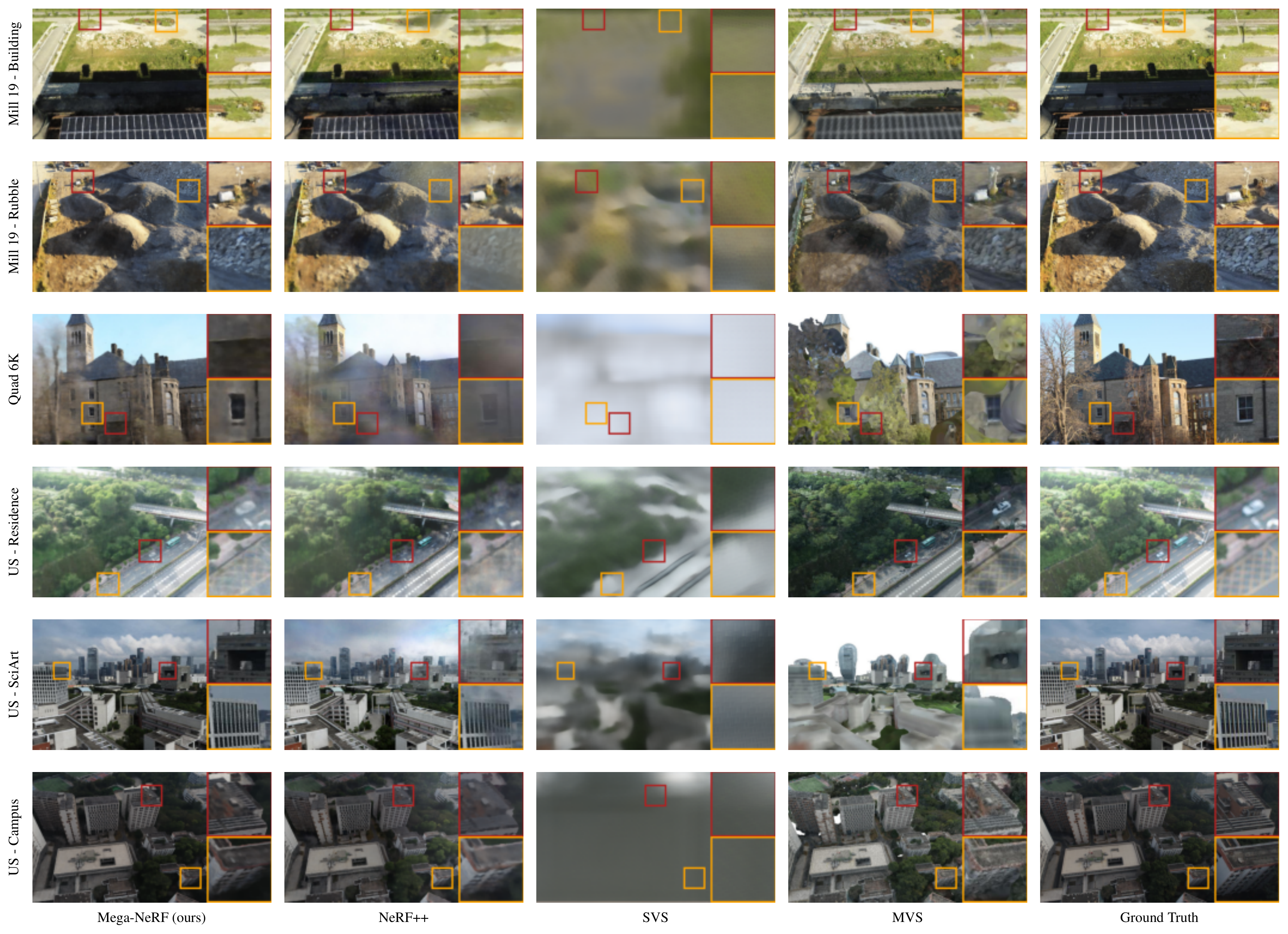}
   \caption{{\bf Scalable training.} \meganerf\ generates the best reconstructions while avoiding the artifacts present in the other approaches.}
\label{fig:qualitative-results}
\vspace*{-5mm}
\end{figure*}

{\bf Qualitative results.} We present two sets of qualitative results. Fig.~\ref{fig:qualitative-results} compares \meganerf's reconstruction quality to existing view synthesis methods. In all cases \meganerf\ captures a high level of detail while avoiding the numerous artifacts present in the other approaches. Fig.~\ref{fig:interactive-qualitative-results} then illustrates the quality of existing fast renderers and our method on top of the same base \meganerf\ model. Our approach generates the highest quality reconstructions in almost all cases, avoiding the pixelization of voxel-based renderers and the blurriness of KiloNeRF.

\subsection{Evaluation protocols}

{\bf Datasets.} We evaluate \meganerf\ against multiple varied datasets. Our \mill\ dataset consists of two scenes we recorded firsthand near a former industrial complex. \mill\ - Building consists of footage captured in a grid pattern across a large $500 \times 250 \ m^2$ area around an industrial building. \mill\ - Rubble covers a nearby construction area full of debris in which we placed human mannequins masquerading as survivors. We also measure \meganerf\ against two publicly available collections - the Quad 6k dataset~\cite{sfm2011cvpr}, a large-scale Structure-from-Motion dataset collected within the Cornell Universty Arts Quad, and several scenes from UrbanScene3D~\cite{UrbanScene3D} which contain high-resolution drone imagery of large-scale urban environments. We refine the initial GPS-derived camera poses in the \mill\ and UrbanScene3D datasets and the estimates provided in the Quad 6k dataset using PixSFM~\cite{lindenberger2021pixsfm}. We use a pretrained semantic segmentation model~\cite{florian2017rethinking} to produce masks of common movable objects in the Quad 6k dataset and ignore masked pixels during training.

{\bf Training.} We evaluate \meganerf\ with 8 submodules each consisting of 8 layers of 256 hidden units and a final fully connected ReLU layer of 128 channels. We use hierarchical sampling during training with 256 coarse and 512 fine samples per ray in the foreground regions and 128/256 samples per ray in the background. In contrast to NeRF, we use the same MLP to query both coarse and fine samples which reduces our model size and allows us to reuse the coarse network outputs during the second rendering stage, saving 25\% model queries per ray. We adopt mixed-precision training to further accelerate the process. We sample 1024 rays per batch and use the Adam optimizer~\cite{adam} with an initial learning rate of $5 \times 10^{-4}$ decaying exponentially to $5 \times 10^{-5}$. We employ the procedure described in \cite{martinbrualla2020nerfw} to finetune \meganerf's appearance embeddings.

\begin{table*}[!htbp]
\resizebox{\textwidth}{!}{
\begin{tabular}{l@{\hspace{1em}}c@{\hspace{1em}}c@{\hspace{1em}}c@{\hspace{1em}}c@{\hspace{2em}}c@{\hspace{1em}}c@{\hspace{1em}}c@{\hspace{1em}}c@{\hspace{2em}}c@{\hspace{1em}}c@{\hspace{1em}}c@{\hspace{1em}}c@{\hspace{2em}}ccc}
\toprule 
&\multicolumn{4}{c}{\mill\ - Building} & \multicolumn{4}{c}{\mill\ - Rubble} & \multicolumn{4}{c}{Quad 6k} \\
& $\uparrow$PSNR & $\uparrow$SSIM & $\downarrow$LPIPS & $\downarrow$Time (h) & $\uparrow$PSNR & $\uparrow$SSIM & $\downarrow$LPIPS & $\downarrow$Time(h) & $\uparrow$PSNR & $\uparrow$SSIM & $\downarrow$LPIPS& $\downarrow$Time(h) \\
NeRF\xspace & 19.54 & 0.525 & 0.512 & 59:51
& 21.14 & 0.522 & 0.546 & 60:21
& 16.75 & 0.559 & 0.616 & 62:48 \\
NeRF++\xspace & 19.48 & 0.520 & 0.514 & 89:02   
& 20.90 & 0.519 & 0.548 & 90:42 
& 16.73 & 0.560 & 0.611 & 90:34 \\
SVS\xspace & 12.59 & 0.299 & 0.778 & 38:17   
& 13.97 & 0.323 & 0.788 & 37:33
& 11.45 & 0.504 & 0.637 & 29:48 \\
DeepView\xspace & 13.28 & 0.295 & 0.751 & 31:20   
& 14.47 & 0.310 & 0.734 & 32:11 
& 11.34 & 0.471 & 0.708 & 19:51 \\
MVS\xspace & 16.45 & 0.451 & 0.545 & 32:29   
& 18.59 & 0.478 & 0.532 & 31:42
& 11.81 & 0.425 & \textbf{0.594} & \textbf{18:55} \\
\meganerf\xspace  & \textbf{20.93} & \textbf{0.547} & \textbf{0.504} & \textbf{29:49}
& \textbf{24.06} & \textbf{0.553} & \textbf{0.516} & \textbf{30:48}
& \textbf{18.13} & \textbf{0.568}  & 0.602 & 39:43 \\
\midrule
&\multicolumn{4}{c}{UrbanScene3D - Residence} & \multicolumn{4}{c}{UrbanScene3D - Sci-Art} & \multicolumn{4}{c}{UrbanScene3D - Campus} \\
& $\uparrow$PSNR & $\uparrow$SSIM & $\downarrow$LPIPS & $\downarrow$Time (h) & $\uparrow$PSNR & $\uparrow$SSIM & $\downarrow$LPIPS & $\downarrow$Time(h) & $\uparrow$PSNR & $\uparrow$SSIM & $\downarrow$LPIPS& $\downarrow$Time(h) \\
NeRF\xspace & 19.01 & 0.593 & 0.488 & 62:40
& 20.70 & 0.727 & 0.418 & 60:15
& 21.83 & 0.521 & 0.630 & 61:56 \\
NeRF++\xspace & 18.99 & 0.586 & 0.493 & 90:48
& 20.83 & 0.755 & 0.393 & 95:00
& 21.81 & 0.520 & 0.630 & 93:50 \\
SVS\xspace & 16.55 & 0.388 & 0.704 & 77:15   
& 15.05 & 0.493 & 0.716 & 59:58 
& 13.45 & 0.356 & 0.773 & 105:01 \\
DeepView\xspace & 13.07 & 0.313 & 0.767 & 30:30
& 12.22 & 0.454 & 0.831 & 31:29
& 13.77 & 0.351 & 0.764 & 33:08 \\
MVS\xspace & 17.18 & 0.532 & \textbf{0.429} & 69:07   
& 14.38 & 0.499 & 0.672 & 73:24
& 16.51 & 0.382 & \textbf{0.581} & 96:01 \\
\meganerf\xspace & \textbf{22.08} & \textbf{0.628} & 0.489 & \textbf{27:20}
& \textbf{25.60} & \textbf{0.770} & \textbf{0.390} & \textbf{27:39} 
& \textbf{23.42} & \textbf{0.537} & 0.618 & \textbf{29:03} \\
\bottomrule
\end{tabular}
}
\caption{{\bf Scalable training.} We compare \meganerf\ to NeRF, NeRF++, Stable View Synthesis (SVS), DeepView, and Multi-View Stereo (MVS) after running each method to completion. \meganerf\ consistently outperforms the baselines even after allowing other approaches to train well beyond 24 hours.}
\vspace*{-5mm}
\label{table:quantitative-results}
\end{table*}

\subsection{Scalable training}
\label{subsection:scalable-training}

{\bf Baselines.} We evaluate \meganerf\ against the original NeRF~\cite{mildenhall2020nerf} architecture and NeRF++~\cite{zhang2020npp}. We also evaluate our approach against Stable View Synthesis~\cite{Riegler2021SVS}, an implementation of DeepView~\cite{flynn2019deepview}, and dense reconstructions from COLMAP~\cite{schoenberger2016mvs}, a traditional Multi-View Stereo approach, as non-neural radiance field-based alternatives.

We use the same Pytorch-based framework and data loading infrastructure across all of NeRF variants to disentangle training speed from implementation specifics. We also use mixed precision training and the same number of samples per ray across all variants. We provide each implementation with the same amount of model capacity as \meganerf\ by setting the MLP width to 2048 units. We provide additional details in Sec.~\ref{sec:baselines} of the supplement.

{\bf Metrics.} We report quantitative results based on PSNR, SSIM~\cite{1284395}, and the VGG implementation of LPIPS~\cite{zhang2018perceptual}. We also report training times as measured on a single machine with 8 V100 GPUs.

{\bf Results.} We run all methods to completion, training all NeRF-based methods for 500,000 iterations. We show results in Table~\ref{table:quantitative-results} along with the time taken to finish training. \meganerf\ outperforms the baselines even after training the other approaches for longer periods.

\begin{figure*}[!htbp]
\centering
\includegraphics[width=\textwidth, clip = true, trim = 0mm 2.7mm 0mm 0mm]{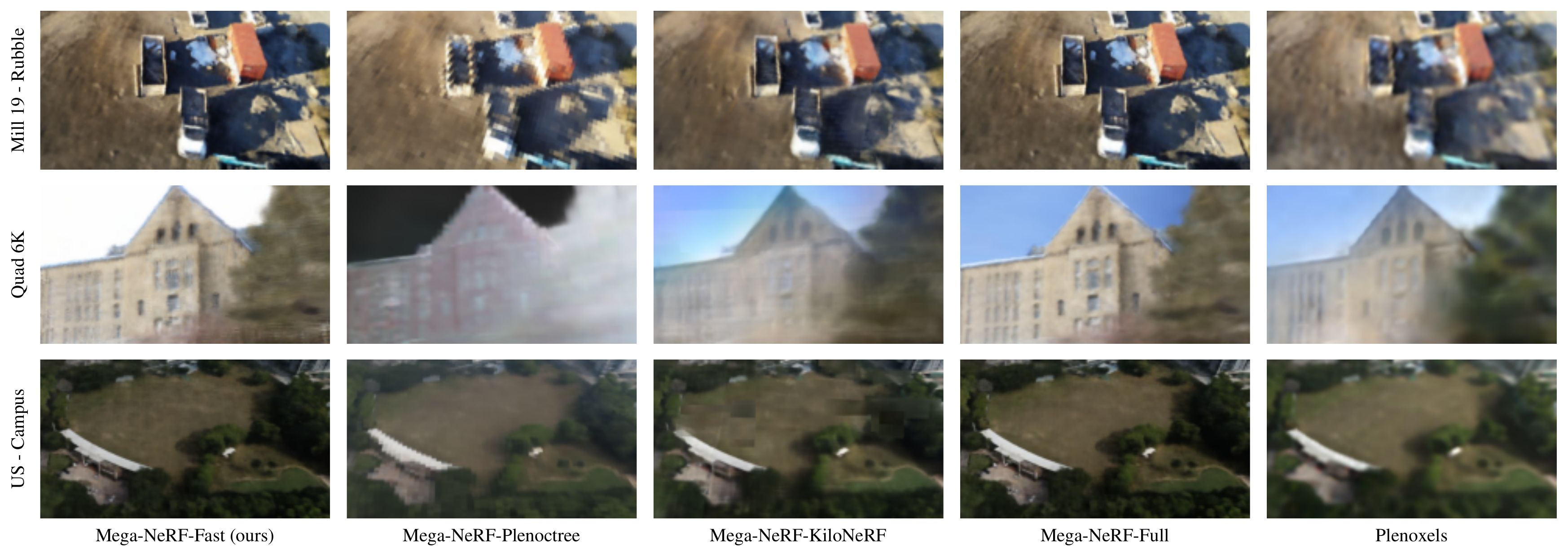}
   \caption{{\bf Interactive rendering.} Plenoctree's approach causes significant voxelization and Plenoxel's renderings are blurry. KiloNeRF's results are crisper but capture less detail than \meganerffast\ and contain numerous visual artifacts.}
\label{fig:interactive-qualitative-results}
\end{figure*}

\begin{table*}[!htbp]
\resizebox{\textwidth}{!}{
\begin{tabular}{l@{\hspace{1em}}c@{\hspace{1em}}c@{\hspace{1em}}c@{\hspace{1em}}c@{\hspace{1em}}c@{\hspace{2em}}c@{\hspace{1em}}c@{\hspace{1em}}c@{\hspace{1em}}c@{\hspace{1em}}c@{\hspace{2em}}c@{\hspace{1em}}c@{\hspace{1em}}c@{\hspace{1em}}c@{\hspace{1em}}c@{\hspace{2em}}cccc}
\toprule 
 \hspace{1em}\textbf{best} \underline{second-best} &\multicolumn{5}{c}{\mill} & \multicolumn{5}{c}{Quad 6k} & \multicolumn{5}{c}{UrbanScene3D} \\
&  &  &  & Preprocess & Render & & &  &  Preprocess &  Render  & &  & & Preprocess & Render \\
&  $\uparrow$PSNR & $\uparrow$SSIM & $\downarrow$LPIPS  & Time (h) & Time (s) & $\uparrow$PSNR & $\uparrow$SSIM & $\downarrow$LPIPS & Time (h) & Time (s) & $\uparrow$PSNR & $\uparrow$SSIM & $\downarrow$LPIPS & Time (h) & Time (s) \\
\meganerf-Plenoctree\xspace & 16.27 & 0.430 & 0.621 & \underline{1:26} & \textbf{0.031}  
& 13.88 & 0.589 & 0.427 & \underline{1:33} & \textbf{0.010}
& 16.41 & 0.498 & 0.530 & \textbf{1:07} & \textbf{0.025} \\
\meganerf-KiloNeRF\xspace & 21.85 & 0.521 & 0.512 & 30:03 & 0.784
& 20.61 & 0.652 & 0.356 & 27:33 & 1.021
& 21.11 & 0.542 & 0.453 & 34:00 & 0.824 \\
\meganerf-Full\xspace & \textbf{22.96} & \textbf{0.588} & \textbf{0.452} & - & 101
& \textbf{21.52} & \textbf{0.676} & \underline{0.355} & - & 174
& \textbf{24.92} & \textbf{0.710} & \textbf{0.393} & - & 122 \\
Plenoxels\xspace & 19.32 & 0.476 & 0.592 & - & 0.482
& 18.61 & 0.645 & 0.411 & - & \underline{0.194}
& 20.06 & 0.608 & 0.503 & - & 0.531 \\

\midrule
\meganerf-Initial\xspace & 17.41 & 0.447 & 0.570 & \textbf{1:08}  & \underline{0.235} 
& 14.30  & 0.585 & 0.386 & \textbf{1:31} & 0.214
& 17.22  & 0.527 & 0.506 & \underline{1:10} & \underline{0.221} \\
\meganerffast\xspace & \underline{22.34} & \underline{0.573} & \underline{0.464} & \textbf{1:08} & 3.96 
& \underline{20.84}  & \underline{0.658} & \textbf{0.342} & \textbf{1:31} & 2.91
& \underline{23.99}  & \underline{0.691} & \underline{0.408} & \underline{1:10} & 3.219 \\
\end{tabular}
}
\caption{
{\bf Interactive rendering.} We evaluate two existing fast renderers on top of our base model, \meganerf-Plenoctree and \meganerf-KiloNeRF, relative to conventional rendering, labeled as \meganerf-Full, Plenoxels, and our novel renderer ({\bf below}). Although PlenOctree achieves a consistently high FPS, its reliance on a finite-resolution voxel structure causes performance to degrade significantly. Our approach remains within 0.8 db in PSNR quality while accelerating rendering by 40x relative to conventional ray sampling.}
\label{table:interactive-quantitative-results}
\end{table*}

\textbf{Diagnostics}. We compare \meganerf\ to several ablations. \meganerf-no-embed removes the appearance embeddings from the model structure. \meganerf-embed-only conversely adds \meganerf's appearance embeddings to the base NeRF architecture. \meganerf-no-bounds uses NeRF++'s unit sphere background/foreground partitioning instead of our formulation described in \ref{subsection:model-architecture}. \meganerf-dense uses fully connected layers instead of spatially-aware sparse connections. \meganerf-joint uses the same model structure as \meganerf\ but trains all submodules jointly using the full dataset instead of using submodule-specific data partitions. We limit training to 24 hours for expediency.

We present our results in Table~\ref{table:ablation-results}. Both the appearance embeddings and the foreground/background decomposition have a significant impact on model performance. \meganerf\ also outperforms both \meganerf-dense and \meganerf-joint, although \meganerf-dense comes close in several scenes. We however note that model sparsity accelerates rendering by 10x relative to fully-connected MLPs and is thus essential for acceptable performance.

\subsection{Interactive exploration}
\label{subsection:interactive-exploration}
{\bf Baselines.} We evaluate two existing fast renderers, Plenoctree and KiloNeRF, in addition to our dynamic renderer. We base all renderers against the same \meganerf\ model trained in \ref{subsection:scalable-training} with the exception of the Plenoctree method which is trained on a variant using spherical harmonics. We accordingly label our rendering variants as \meganerf-Plenoctree, \meganerf-KiloNeRF, and \meganerffast\ respectively. We measure traditional NeRF rendering as an additional baseline, which we refer to as \meganerf-Full, and Plenoxels~\cite{yu2022plenoxels} which generates a sparse voxel structure similar to Plenoctree but with trilinear instead of nearest-neighbor interpolation.

{\bf Metrics.} We report the same perceptual metrics as in \ref{subsection:scalable-training} and the time it takes to render a 720p image. We evaluate only foreground regions as Plenoctree and KiloNeRF assume bounded scenes. We also report any additional time needed to generate any additional data structures needed for rendering \textit{beyond} the base model training time in the spirit of enabling fly-throughs within a day. As our renderer presents an initial coarse voxel-based estimate before progressively refining the image, we present an additional set of measurements, labeled as \meganerf-Initial, to quantify the quality and latency of the initial reconstruction.

\begin{table*}[!htbp]
\resizebox{\textwidth}{!}{
\begin{tabular}{l@{\hspace{1em}}c@{\hspace{1em}}c@{\hspace{1em}}c@{\hspace{2em}}c@{\hspace{1em}}c@{\hspace{1em}}c@{\hspace{2em}}c@{\hspace{1em}}c@{\hspace{1em}}c@{\hspace{2em}}ccc}
\toprule 
&\multicolumn{3}{c}{\mill} & \multicolumn{3}{c}{Quad 6k} & \multicolumn{3}{c}{UrbanScene3D} \\
& $\uparrow$PSNR & $\uparrow$SSIM & $\downarrow$LPIPS & $\uparrow$PSNR & $\uparrow$SSIM & $\downarrow$LPIPS & $\uparrow$PSNR & $\uparrow$SSIM & $\downarrow$LPIPS \\
\meganerf-no-embed\xspace & 20.42 & 0.500 & 0.561 
& 16.16 & 0.544 & 0.643
& 19.45 & 0.587 & 0.545 \\ 
\meganerf-embed-only\xspace & 21.48 & 0.494 & 0.566 
& 17.91 & 0.559 & 0.638
& 22.79 & 0.611 & 0.537 \\ 
\meganerf-no-bounds\xspace & 22.14 & 0.534 & 0.522   
& 18.02 & 0.565 & 0.616
& 23.42 & 0.636 & 0.511 \\
\meganerf-dense\xspace & 21.63 & 0.504 & 0.551   
& 17.94 & 0.562 & 0.627
& 22.44 & 0.605 & 0.558 \\
\meganerf-joint\xspace & 21.10 & 0.490 & 0.574    
& 17.43 & 0.560 & 0.616
& 21.45 & 0.595 & 0.567 \\ 
\meganerf\xspace  & \textbf{22.34} & \textbf{0.540} & \textbf{0.518}
& \textbf{18.08} & \textbf{0.566} & \textbf{0.602}
& \textbf{23.60} & \textbf{0.641} & \textbf{0.504} \\   
\bottomrule
\end{tabular}
}
\caption{\textbf{Diagnostics.} We compare \meganerf\ to various ablations after 24 hours of training. Each individual component contributes significantly to overall model performance.}
\vspace*{-3mm}
\label{table:ablation-results}
\end{table*}

{\bf Results.} We list our results in Table~\ref{table:interactive-quantitative-results}. Although \meganerf-Plenoctree renders most quickly, voxelization has a large visual impact. Plenoxels provides better renderings but still suffers from the same finite resolution shortfalls and is blurry relative to the NeRF-based methods. \meganerf-KiloNeRF comes close to interactivity at 1.1 FPS but still suffers from noticeable visual artifacts. Its knowledge distillation and finetuning processes also require over a day of additional processing. In contrast, \meganerffast\ remains within 0.8 db in PSNR of normal NeRF rendering while providing a 40x speedup. \meganerf-Plenoctree and \meganerffast\ both take an hour to build similar octree structures.

\section{Limitations}

We discuss limitations and the societal impact of our work in the supplementary material.

\section{Conclusion}

We present a modular approach for building NeRFs at previously unexplored scale. We introduce a sparse and spatially aware network structure along with a simple geometric clustering algorithm that partitions training pixels into different NeRF submodules which can be trained in parallel. These modifications speed up training by over 3x while significantly improving reconstruction quality. Our empirical evaluation of existing fast renderers on top of \meganerf\ suggests that interactive NeRF-based rendering at scale remains an open research question. We advocate leveraging temporal smoothness to minimize redundant computation between views as a valuable first step.

\section*{Acknowledgments}

{\footnotesize
  This research was supported by the National Science Foundation (NSF) under grant number CNS-2106862, the Defense Science and Technology Agency of Singapore (DSTA), and the CMU Argo AI Center for Autonomous Vehicle Research. Additional support was provided by CableLabs, Crown Castle, Deutsche Telekom, Intel, InterDigital, Microsoft, Seagate, VMware, Vodafone, and the Conklin Kistler family fund.  Any opinions, findings, conclusions or recommendations expressed in this material are those of the authors and do not necessarily reflect the view(s) of their employers or the above funding sources.
\par}

{\small
\bibliographystyle{ieee_fullname}
\bibliography{main}
}

\clearpage
\appendix

\section*{Supplemental Materials}

\section{Data Pruning}
\label{sec:data-pruning}

\begin{figure}
\centering
\includegraphics[width=\linewidth, clip = true, trim = 0mm 0mm 0mm 0mm]{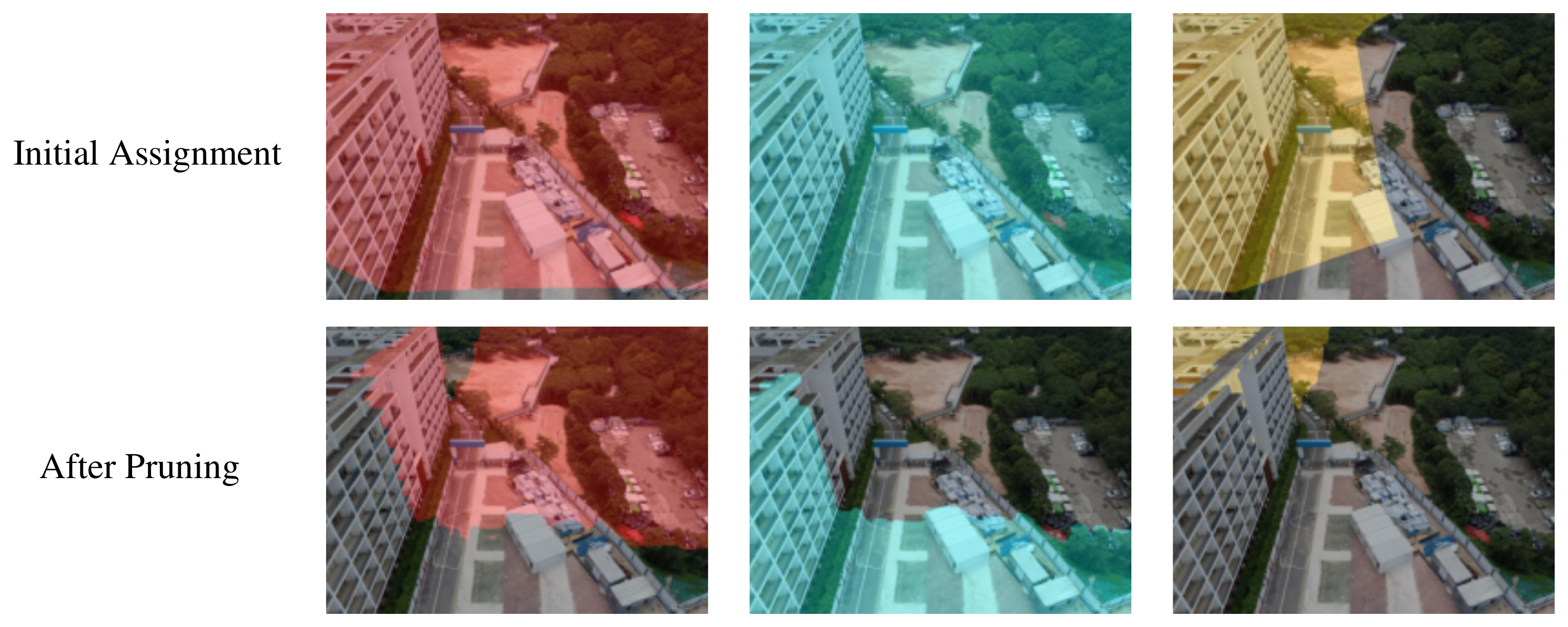}
   \caption{{\bf Data pruning.} The initial assignment of pixels to cells is based purely on camera positions. We add each pixel to the training set of \textbf{all} cells it traverses, leading to overlap between sets \textbf{(top)}. After the model gains a 3D understanding of the scene, we can filter irrelevant pixels by instead assigning pixels based on camera ray intersection with solid surfaces \textbf{(bottom)}.}
   \vspace*{-5mm}
\label{fig:data-pruning}
\end{figure}

Recall that the initial assignment of pixels to spatial cells is based on camera positions, irrespective of scene geometry (because that is not known at initialization time). However, Sec.~\ref{subsection:training} points out that one could repartition our training sets with additional 3D knowledge. Intuitively, one can prune away irrelevant pixel/ray assignments that don't contribute to a particular NeRF submodule due to an intervening occluder (Fig.~\ref{fig:data-pruning}).

To explore this optimization, we further prune each data partition early into the training process after the model gains a coarse 3D understanding of the scene (100,000 iterations in our experiments). As directly querying depth information using conventional NeRF rendering is prohibitive at our scale, we instead take inspiration from Plenoctree and tabulate the scene's model opacity values into a fixed resolution structure. We then calculate the intersection of each training pixel's camera ray against surfaces within the structure to generate new assignments. We found that it took around 10 minutes to compute the model density values and 500ms per image to generate the new assignments. We summarize our findings in Table~\ref{table:deduplicated}.

\begin{table*}[!htbp]
\resizebox{\textwidth}{!}{
\begin{tabular}{l@{\hspace{1em}}c@{\hspace{1em}}c@{\hspace{1em}}c@{\hspace{1em}}c@{\hspace{2em}}c@{\hspace{1em}}c@{\hspace{1em}}c@{\hspace{1em}}c@{\hspace{2em}}c@{\hspace{1em}}c@{\hspace{1em}}c@{\hspace{1em}}c@{\hspace{2em}}cccc}
\toprule 
&\multicolumn{4}{c}{\mill} & \multicolumn{4}{c}{Quad 6k} & \multicolumn{4}{c}{UrbanScene3D} \\
&  $\uparrow$PSNR & $\uparrow$SSIM & $\downarrow$LPIPS & $\downarrow$Pixels & $\uparrow$PSNR & $\uparrow$SSIM & $\downarrow$LPIPS & $\downarrow$Pixels & $\uparrow$PSNR & $\uparrow$SSIM & $\downarrow$LPIPS & $\downarrow$Pixels \\
Original Data\xspace & 22.50 & 0.550 & 0.511 & 0.211
& 18.13 & 0.568 & 0.602 & 0.390
& 23.65 & 0.644 & 0.500 & 0.270 \\
Pruned Data\xspace & \textbf{22.76} & \textbf{0.571} & \textbf{0.488} & \textbf{0.160}
& \textbf{18.16} & \textbf{0.569} & \textbf{0.593} & \textbf{0.149}
& \textbf{23.87} & \textbf{0.656} & \textbf{0.483} & \textbf{0.163} \\
\bottomrule 
\end{tabular}
}
\caption{{\bf Data pruning.} The initial assignment of pixels to spatial cells is based purely on rays emanating from camera centers, irrespective of scene geometry. However, once a rough \meganerf\ has been trained, coarse estimates of scene geometry can be used to prune irrelevant pixel assignments. Doing so reduces the amount of training data for each submodule by up to 2x while increasing accuracy for a fixed number of 500,000 iterations.
}
\label{table:deduplicated}
\end{table*}

\section{Scaling properties}
\label{sec:scaling-properties}

\begin{figure*}[!htbp]
    \includegraphics[width=\textwidth, clip = true, trim = 0mm 3.5mm 0mm 0mm]{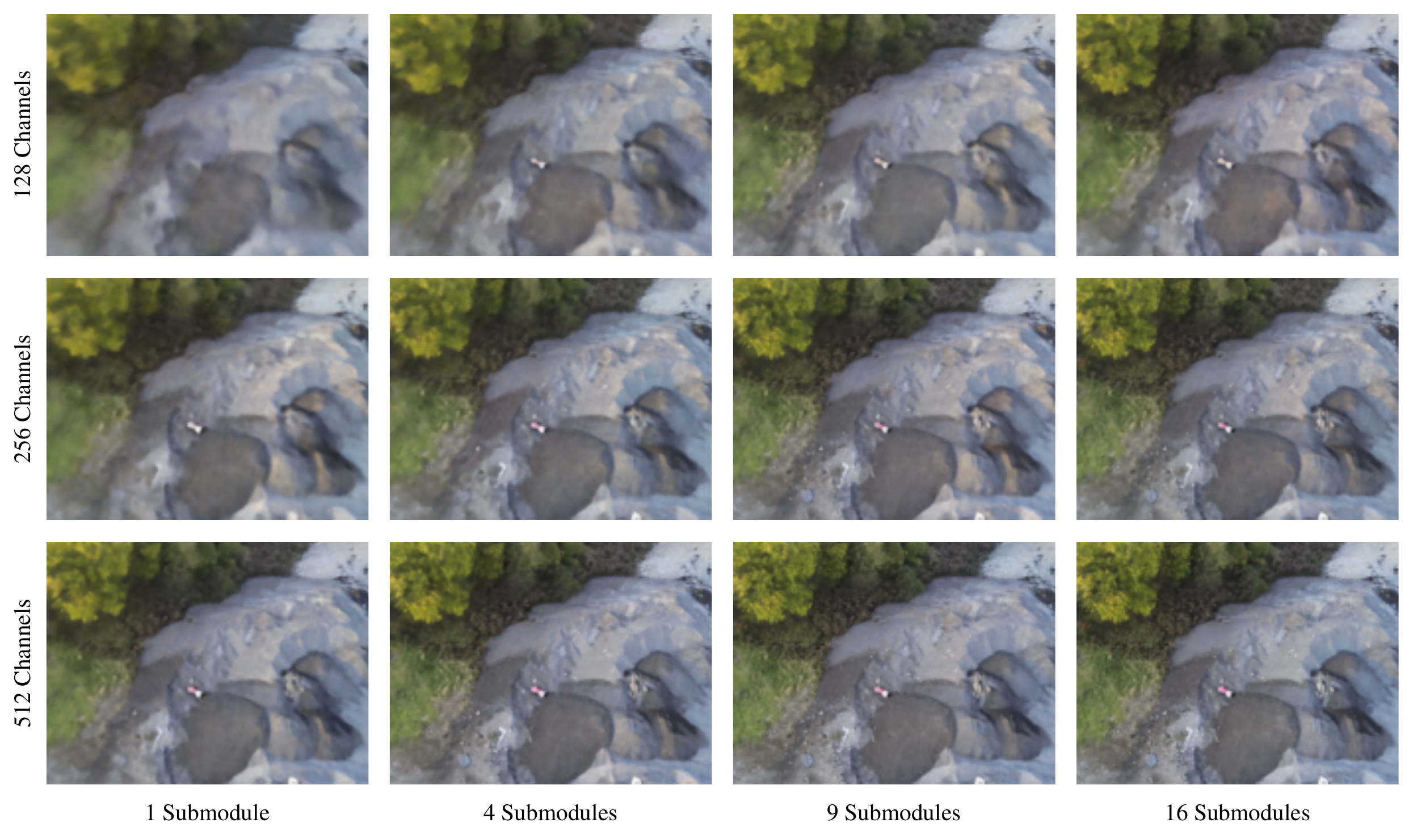}
    \caption{\textbf{Model scaling.} Example rendering within our \mill\ - Rubble dataset across different numbers of submodules ({\bf columns}) and channels per submodule ({\bf rows}). \meganerf\ generates increasingly photo-realistic renderings as capacity increases. Increasing the number of submodules increases the overall model capacity with little impact to training and inference time.}
\label{fig:channels-and-cells-qualitative}
\vspace*{-5mm}
\end{figure*}

We further explore \meganerf's scaling properties against the \mill\ - Rubble dataset. We vary the total number of submodules and the number of channels per submodule across 1, 4, 9, and 16 submodules and 128, 256, and 512 channels respectively. We summarize our findings in Table~\ref{table:scaling-by-channels}. Increasing the model capacity along either dimension improves rendering quality, as depicted in Fig.~\ref{fig:channels-and-cells-qualitative}. However, although increasing the channel count severely penalizes training and rendering speed, the number of submodules has less impact.

\begin{table*}[!htbp]
\resizebox{\textwidth}{!}{
\begin{tabular}{l@{\hspace{1em}}c@{\hspace{1em}}c@{\hspace{1em}}c@{\hspace{1em}}c@{\hspace{1em}}c@{\hspace{2em}}c@{\hspace{1em}}c@{\hspace{1em}}c@{\hspace{1em}}c@{\hspace{1em}}c@{\hspace{2em}}c@{\hspace{1em}}c@{\hspace{1em}}c@{\hspace{1em}}c@{\hspace{1em}}c@{\hspace{2em}}c@{\hspace{1em}}c@{\hspace{1em}}c@{\hspace{1em}}c@{\hspace{1em}}c@{\hspace{2em}}cccc}
\toprule 
&\multicolumn{5}{c}{1 Submodule} & \multicolumn{5}{c}{4 Submodules} & \multicolumn{5}{c}{9 Submodules} & \multicolumn{5}{c}{16 Submodules} \\
&  &  & & Train & Render &  &  &  & Train & Render &  &  & & Train & Render & $\uparrow$PSNR & $\uparrow$SSIM & & Train  & Render \\
&  $\uparrow$PSNR & $\uparrow$SSIM & $\downarrow$LPIPS & Time (h) & Time (s) & $\uparrow$PSNR & $\uparrow$SSIM & $\downarrow$LPIPS & Time (h) & Time (s) & $\uparrow$PSNR & $\uparrow$SSIM & $\downarrow$LPIPS & Time (h) & Time (s) & $\uparrow$PSNR & $\uparrow$SSIM & $\downarrow$LPIPS & Time (h) & Time (s)\\
128 Channels\xspace & 21.75 & 0.435 & 0.670 & \textbf{18:54} & \textbf{2.154}
& 22.61 & 0.469 & 0.631 & \textbf{18:56} & \textbf{2.489}
& 23.08 & 0.495 & 0.594 & \textbf{19:01} & \textbf{2.633}
& 23.34 & 0.513 & 0.568 & \textbf{19:02} & \textbf{2.851} \\
256 Channels\xspace & 22.60 & 0.471 & 0.622 & 28:54 & 3.298
& 23.63 & 0.521 & 0.551 & 29:09 & 3.427
& 24.17 & 0.559 & 0.508 & 29:13 & 3.793
& 24.52 & 0.584 & 0.481 & 29:14 & 3.991 \\
512 Channels\xspace & \textbf{23.40} & \textbf{0.512} & \textbf{0.559} & 52:33 & 6.195
& \textbf{24.53} & \textbf{0.581} & \textbf{0.482} & 52:34 & 6.313
& \textbf{25.11} & \textbf{0.625} & \textbf{0.438} & 53:36 & 6.671
& \textbf{25.68} & \textbf{0.659} & \textbf{0.407} & 53:45 & 6.870 \\
\bottomrule 
\end{tabular}
}
\caption{\textbf{Model scaling.} We scale up \meganerf\ with additional submodules ({\bf rows}) and increased channel count per submodule ({\bf columns}). Scaling up both increases reconstruction quality, but increasing channels significantly increases both training and rendering time (as measured for \meganerffast).}
\label{table:scaling-by-channels}
\end{table*}

\section{Dynamic octree generation}
\label{sec:octree-generation}

The maximum tree size used by \meganerf-Dynamic is bounded by available GPU memory, which we set to 20M elements in our experiments. We track the number of pixels visible from each node as we traverse the tree when rendering. We then subdivide the top $k$ (16,384) nodes with the most pixels. We observe maximum tree depths of roughly 12 in practice. As we track which nodes contribute to which pixels, we also prune entries that have not recently contributed in order to reclaim space whenever we hit capacity.

\section{Baselines}
\label{sec:baselines}

\textbf{Multi-view stereo.} Although scaling dense multi-view stereo remains an open research problem~\cite{5539802}, we optimize for the best possible reconstructions instead of training time for the purpose of our evaluation. We use COLMAP to generate meshes with Poisson surface reconstruction. We found that this method failed to generate reasonable results for scenes containing many sky pixels. We therefore use foreground masks generated from a trained \meganerf\ model to mask background regions during surface reconstruction to achieve better results.

\textbf{Stable View Synthesis.} We train the network representing each scene from scratch for 600,000 iterations. Stable View Synthesis relies on a geometric scaffold containing depth information and we use the meshes generated from COLMAP for this purpose.

\textbf{DeepView.} We base our DeepView baseline on a publicly available implementation. We use 3 blocks of 24 channels and train our model for 200,000 iterations using random 200 x 100 crops of the input views. During training, we randomly sample nearby input views for a given target view as determined by the capture time of each image.

\section{Limitations}

\textbf{Pose accuracy.} Although our work presents a first step towards scaling NeRF to handle large-scale view synthesis, several obstacles remain ahead of deploying them in practice. Pose accuracy is arguably the largest limiting factor. The initial models we trained using raw camera poses collected from standard drone GPS and IMU sensors were extremely blurry. As alternatives to PixSFM~\cite{lindenberger2021pixsfm}, we experimented with refining our camera poses with BARF's~\cite{lin2021barf} coarse-to-fine adjustment and Pix4DMapper~\cite{pix4d-mapper}, a commercial drone mapping solution. Our results were uniformly better with the PixSFM poses, with a PSNR gap of over 6 db relative to the second-best solution (Pix4DMapper). SCNeRF~\cite{SCNeRF2021} and GNeRF~\cite{meng2021gnerf} are other recent alternatives that merit further exploration. Another hardware-based solution would be to use higher-accuracy RTK GPS modules when collecting footage.

\textbf{Dynamic objects.} We did not explicitly address dynamic scenes within our work, a relevant factor for many human-centered use cases. Several recent NeRF-related efforts, including NR-NeRF~\cite{tretschk2021nonrigid}, Nerfies~\cite{park2021nerfies}, NeRFlow~\cite{du2021nerflow}, and DynamicMVS~\cite{Gao-freeviewvideo} focus on dynamism, but we theorize that scaling these approaches to larger urban scenes will require additional work.

\textbf{Scale.} \meganerf\ explicitly targets urban-scale environments instead of smaller single-object settings. Our tests against scenes from the Synthetic-NeRF dataset suggests our ray bound strategy and per-image appearance embeddings do not harm quality but that our spatial partitioning strategy reduces PSNR by about 1 db relative to NeRF.

\textbf{Rendering speed.} While our renderer avoids the pitfalls of existing fast NeRF approaches, it does not quite reach the throughput needed for truly interactive applications. We explored uncertainty-based methods as detailed in~\cite{DBLP:journals/corr/abs-2105-09103} to further improve sampling efficiency, but open challenges remain. 

\textbf{Training speed.} Although our training process is several factors quicker than previous works, NeRF training time remains a significant bottleneck towards rapid model deployment. Recent methods such as pixelNeRF~\cite{9577688} and GRF~\cite{grf2020} that introduce conditional priors would likely complement our efforts but necessitate gathering similar data to the scenes that we target. We hope that our \mill\ dataset, in addition to existing collections such as UrbanScene3D~\cite{UrbanScene3D}, will serve as a valuable contribution.

\section{Societal impact}

The capture of drone footage brings with it the possibility of inadvertently and accidentally capturing privacy-sensitive information such as people's faces and vehicle license plate numbers. Furthermore, what is considered sensitive and not can vary widely depending on the context.

We are exploring the technique of ``denaturing" first described by Wang et al~\cite{10.1145/3083187.3083192} that allows for fine-grain policy guided removal of sensitive pixels at interactive frame rates. As denaturing can be done at full frame rate, preprocessing should not slow down training, although it is unclear what the impact of the altered pixels would have on the resulting model. We plan on investigating this further in the future. 

\begin{figure}
    \includegraphics[width=\linewidth]{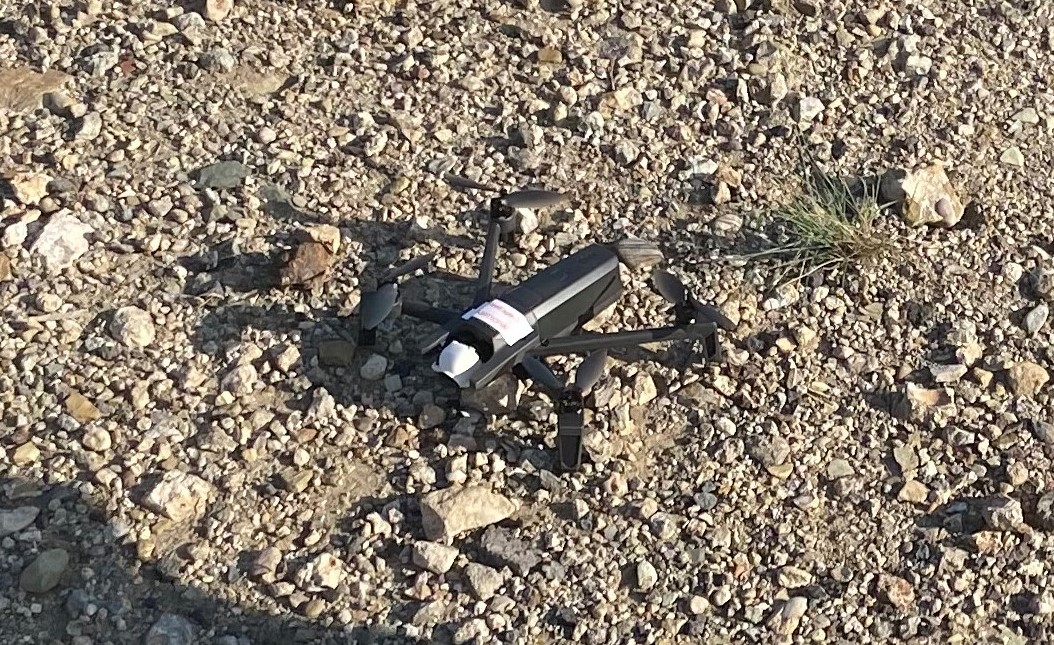}
    \caption{{\bf Parrot ANAFI drone}. The drone used to collect data for the \mill\ dataset. The drone comes equipped with a 4K Camera, GPS, and an inertial measurement unit (IMU) from which we derive initial camera poses.}
\label{fig:anafi-1}
\vspace*{-5mm}
\end{figure}

\begin{figure}
    \includegraphics[width=\linewidth]{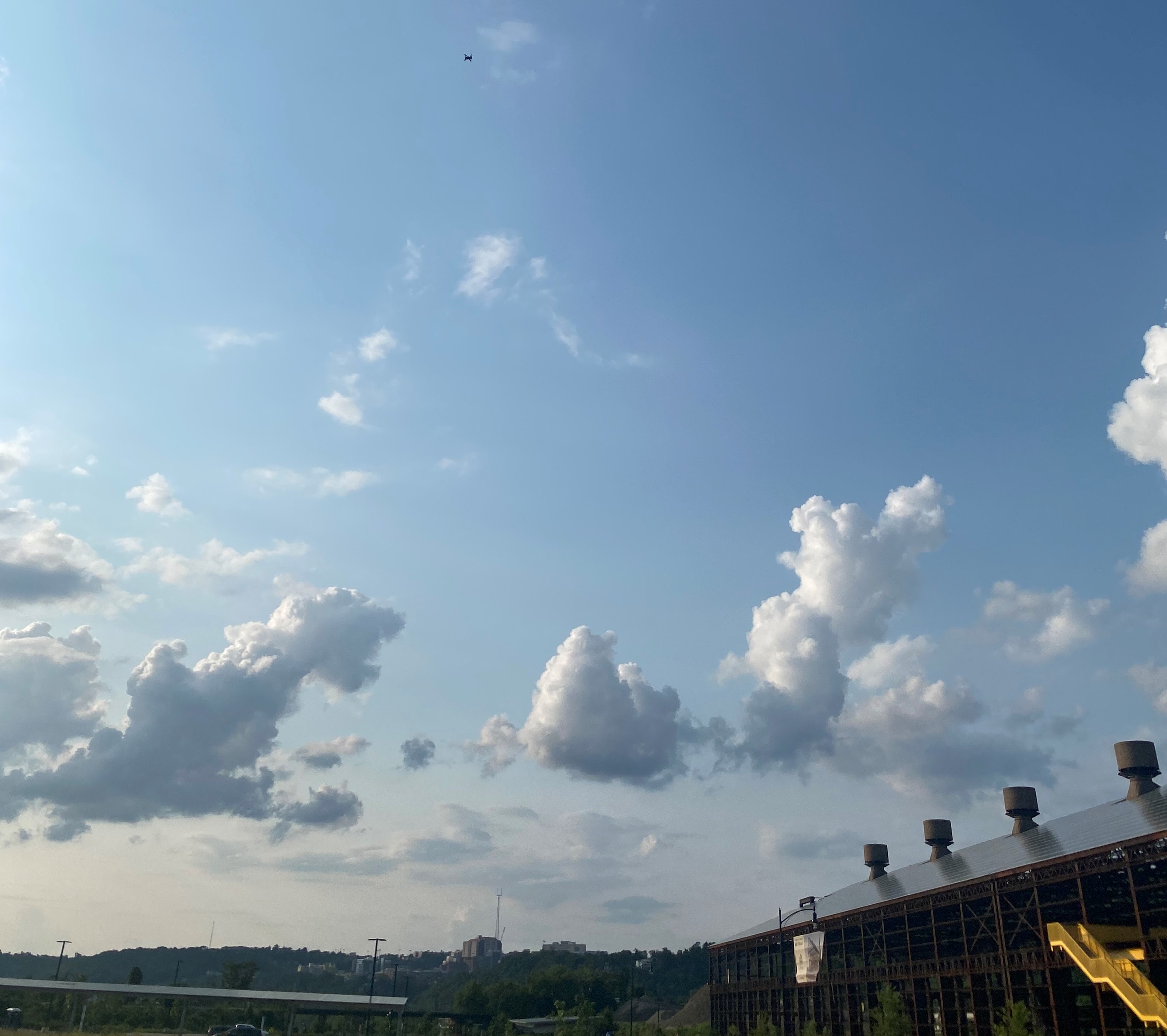}
    \caption{Our drone flying above the \mill\ - Building scene.}
\label{fig:anafi-2}
\end{figure}

\section{Assets}

\begin{figure}
    \includegraphics[width=\linewidth]{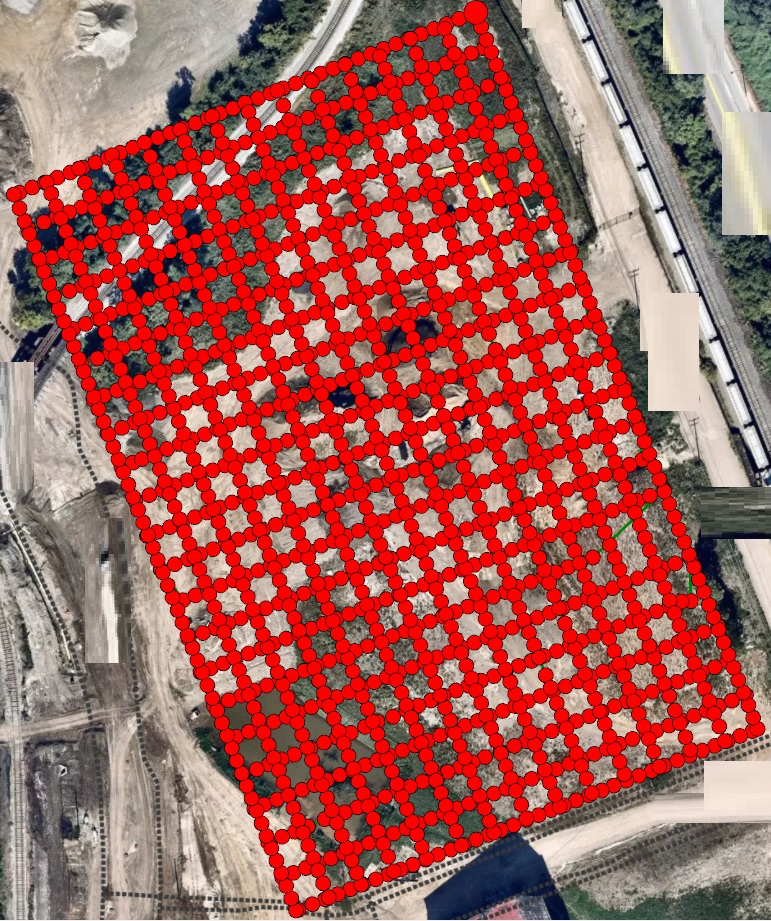}
    \caption{{\bf Data collection}. We collect our poses for our \mill\ dataset using a grid pattern as shown. Collecting footage across a 200,000 $m^2$ area takes approximately 2 hours.}
\label{fig:rubble-poses}
\vspace*{-5mm}
\end{figure}

\textbf{\mill\ dataset.} We have publicly released our \mill\ dataset along with our calibrated poses to the wider research community. We collected our data using the Parrot ANAFI drone pictured in Figs.~\ref{fig:anafi-1} and \ref{fig:anafi-2}. We captured photos in the grid pattern illustrated in Fig.~\ref{fig:rubble-poses}. In addition to FAA approval, we also received permission from the city to fly in the area and ensured that no non-consenting people were captured in the data.

\begin{figure*}[!htbp]
\centering
\includegraphics[width=\textwidth, clip = true, trim = 0mm 2.7mm 0mm 0mm]{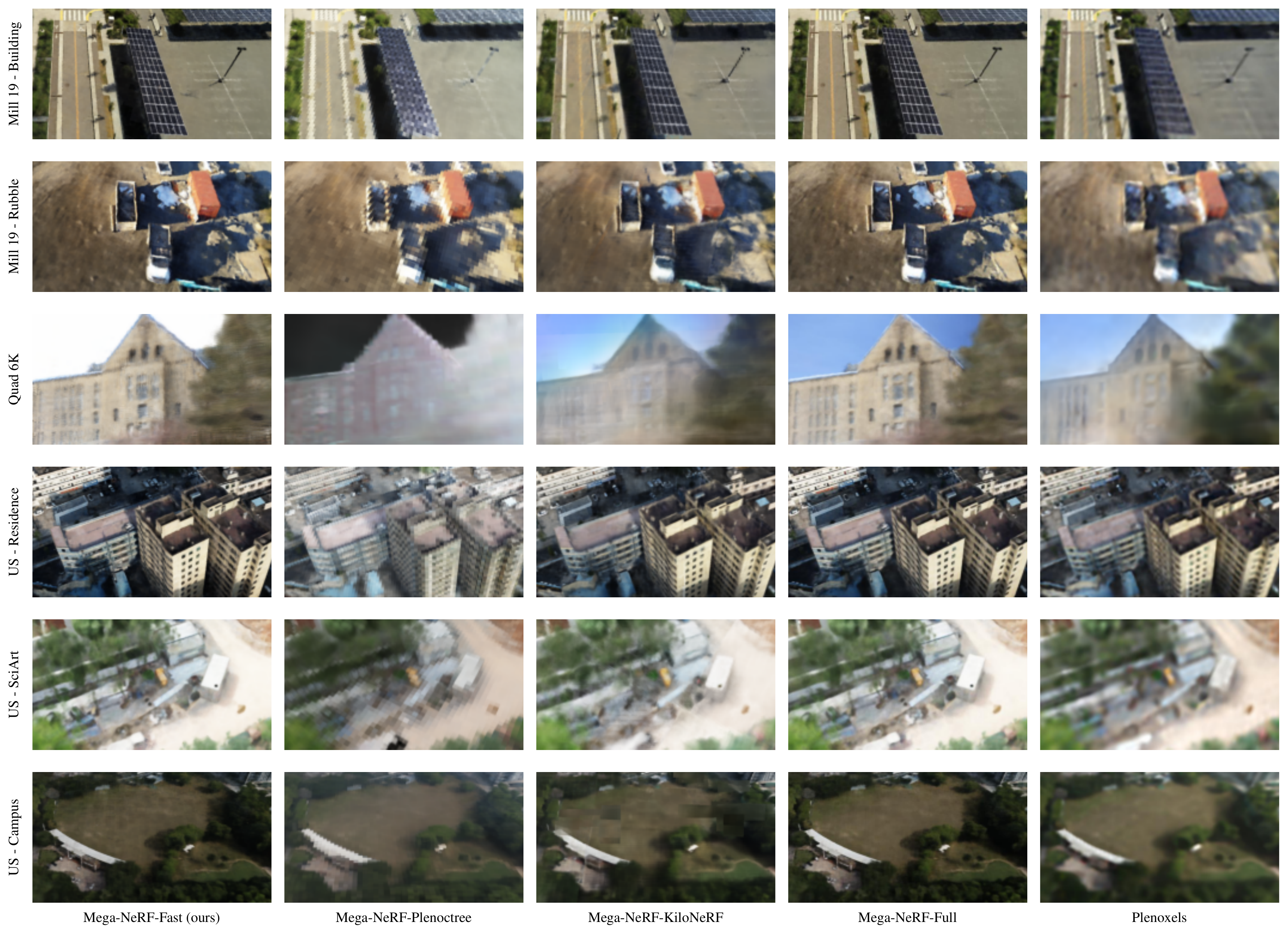}
   \caption{Additional interactive rendering results.}
\label{fig:interactive-qualitative-results-full}
\end{figure*}

\textbf{Third-party assets.} The main third-party assets we used were the UrbanScene3D~\cite{UrbanScene3D} dataset, the Quad 6k dataset~\cite{sfm2011cvpr}, and Pytorch~\cite{NEURIPS2019_9015}, all of which are cited in our main paper. Pytorch uses a BSD license and the UrbanScene3D dataset is freely available for research and education use only. The Quad 6k dataset does not include an explicit license but is freely available at ~\url{http://vision.soic.indiana.edu/projects/disco/}.

\section{Dataset statistics}
\label{sec:more-statistics}

\textbf{Visibility.} We generate the visibility statistics in Table~\ref{table:dataset-statistics} by first training a NeRF model for each scene. As in Sec.~\ref{sec:data-pruning}, we compute and store opacity values into a fixed resolution structure and project camera rays for all images in the scene. We then measure the proportion of surface voxels each image's rays intersects relative to the total number in the scene. 

\textbf{Additional datasets.} We provide top-level statistics for commonly used view synthesis datasets in Table~\ref{table:additional-dataset-statistics} to complement those in Table~\ref{table:dataset-statistics}.

\begin{table}
\resizebox{\linewidth}{!}{
\begin{tabular}{l@{\hspace{1em}}c@{\hspace{1em}}c@{\hspace{1em}}c@{\hspace{1em}}c}
\toprule 
&  Resolution & \# Images & \# Pixels/Rays \\
Synthetic NeRF~\cite{mildenhall2020nerf} \xspace & 400 x 400 & 400 & 256,000,000  \\
LLFF~\cite{mildenhall2019llff} \xspace & 4032 x 3024 & 41 & 496,419,840 \\
Light Field~\cite{Yucer:LightfieldSegmentation:2016} \xspace & 1280 x 720 & 214 & 195,910,200 \\
Tanks and Temples~\cite{Knapitsch2017}\xspace & 1920 x 1080 & 283 & 587,658,240 \\
Phototourism~\cite{phototourism}\xspace & 919 x 794 & 1708 & 1,149,113,846 \\
\midrule
\mill\xspace     & 4608 x 3456 & 1809 & 28,808,773,632 \\
Quad 6k~\cite{sfm2011cvpr}\xspace   & 1708 x 1329 & 5147 & 11,574,265,679 \\
UrbanScene3D~\cite{UrbanScene3D}\xspace   & 5232 x 3648 & 3824 & 74,102,106,112 \\
\bottomrule
\end{tabular}
}
\caption{Comparison of datasets commonly used in view synthesis ({\bf above}) relative to those evaluated in our work ({\bf below}). We average the resolution, number of images, and total number of pixels across each captured scene. We report statistics for Light Field and Tanks and Temples using the splits in \cite{zhang2020npp} and \cite{yu2021plenoctrees} respectively. For Phototourism we average across the scenes used in \cite{martinbrualla2020nerfw}.}
\label{table:additional-dataset-statistics}
\end{table}

\section{Additional results}

We include additional interactive rendering results across all datasets in Fig.~\ref{fig:interactive-qualitative-results-full} to complement those in Fig.~\ref{fig:interactive-qualitative-results}.

\end{document}